\RequirePackage{fix-cm}
\documentclass[twocolumn]{svjour3}          
\smartqed  
\usepackage{graphicx}
\usepackage{algorithm}
\usepackage{algpseudocode}
\usepackage{amsmath} 
\usepackage{amssymb}  
\usepackage{amsbsy}
\usepackage{subcaption}
\usepackage{rotating}
\usepackage{relsize}

\usepackage{verbatim}
\usepackage{todonotes}
\usepackage{booktabs}
\usepackage{color}
\usepackage{bbding}
\usepackage{dirtytalk}
\usepackage{xcolor}

\usepackage{tikz}
\usetikzlibrary{fit,calc}
\newcommand*{\tikzmk}[1]{\tikz[remember picture,overlay,] \node (#1) {};\ignorespaces}
\newcommand{\boxita}[1]{\tikz[remember picture,overlay]{\node[xshift=-1.25cm,yshift=-10pt,fill=#1,opacity=.25,fit={(A)($(B)+(1\linewidth,1\baselineskip)$)}] {};}\ignorespaces}
\newcommand{\boxitb}[1]{\tikz[remember picture,overlay]{\node[xshift=-2.7cm,yshift=-10pt,fill=#1,opacity=.25,fit={(B)($(C)+(1.182\linewidth,1\baselineskip)$)}] {};}\ignorespaces}
\newcommand{\boxitc}[1]{\tikz[remember picture,overlay]{\node[xshift=-4.83cm,yshift=-10pt,fill=#1,opacity=.25,fit={(C)($(D)+(1.453\linewidth,1\baselineskip)$)}] {};}\ignorespaces}
\colorlet{pink1}{red!40}
\colorlet{blue1}{cyan!60}
\colorlet{green1}{green!40}

\usepackage{enumerate}

\newcommand{\change}[1]{\textcolor{black}{#1}}
\newcommand{\changetwo}[1]{\textcolor{black}{#1}}
\newcommand{\changethree}[1]{\textcolor{black}{#1}}
\newcommand{\eg} {\textit{e.g.,}~} %
\newcommand{\etal}{\MakeLowercase{\textit{et al.\ }}}

\usepackage{soul}
\providecommand  {\fa}   [1]{{\color{black}#1}}

 \journalname{Autonomous Robots}
\begin{document}

\title{\changetwo{Imitation learning-based framework for learning 6-D linear compliant motions}
\thanks{This work was supported by Academy of Finland, decision 286580.}
}


\author{Markku Suomalainen         \and
		Fares J. Abu-Dakka         \and
        Ville Kyrki 
}


\institute{Markku Suomalainen \at
              University of Oulu \& Aalto University \\
              Tel.: +358 50 5286 116\\
              \email{markku.suomalainen@oulu.fi}           
           \and
           Fares J. Abu-Dakka \at
              Intelligent Robotics Group, Department of Electrical Engineering and Automation (EEA), Aalto University, Espoo, Finland. \\ 
              \email{fares.abu-dakka@aalto.fi}
           \and
           Ville Kyrki \at
              Intelligent Robotics Group, Department of Electrical Engineering and Automation (EEA), Aalto University, Espoo, Finland. \\ 
              Tel.: +358 50 408 2035 \\
              \email{ville.kyrki@aalto.fi}
}

\date{Received: date / Accepted: date}

\maketitle

\begin{abstract}
We present a novel method for learning from demonstration 6-D tasks that can be modeled as a sequence of linear motions and compliances. The focus of this paper is the learning of a single linear primitive, many of which can be sequenced to perform more complex tasks. \changethree{The presented method learns from demonstrations how to take advantage of mechanical gradients in in-contact tasks, such as assembly, both for translations and rotations, without any prior information}. The method assumes there exists a desired linear direction in 6-D which, if followed by the manipulator, leads the robot's end-effector to the goal area shown in the demonstration, either in free space or by leveraging contact through compliance. First, demonstrations are gathered where the teacher explicitly shows the robot how the mechanical gradients can be used as guidance towards the goal. From the demonstrations, a set of directions is computed which would result in the observed motion at each timestep during a demonstration of a single primitive. By observing which direction is included in all these sets, we find a single desired direction which can reproduce the demonstrated motion. Finding the number of compliant axes and their directions in both rotation and translation is based on the assumption that in the presence of a desired direction of motion, all other observed motion is caused by the contact force of the environment, signalling the need for compliance. We evaluate the method on a KUKA LWR4+ robot with test setups imitating typical tasks where a human would use compliance to cope with positional uncertainty. Results show that the method can successfully learn and reproduce compliant motions by taking advantage of the geometry of the task, therefore reducing the need for localization accuracy.
\end{abstract}

\section{INTRODUCTION}
\label{intro}
Currently industrial robots are often confined inside mass production factories, where the environment can be precisely modelled and controlled and the production batch sizes are high. However, the use of robots is steadily rising and they are expected to take over households, construction yards and factories within the near future. \changetwo{There is tremendous potential for fast and efficiently automatization of tasks that are recurring frequently but in smaller batches than in car factories.}

\begin{figure}[t]
\centering
\begin{subfigure}[b]{0.9\columnwidth}
	\centering
	\includegraphics[width=\columnwidth]{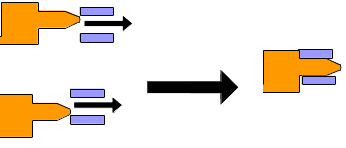}
	\caption{Position alignment.}
	\label{fig:assembly}
\end{subfigure}
\begin{subfigure}[b]{0.6\columnwidth}
	\centering
	\includegraphics[width=\columnwidth]{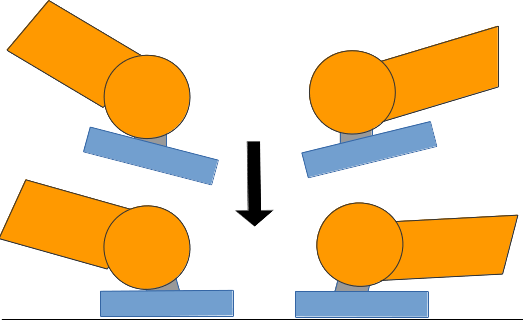}
	\caption{Orientation alignment}
	\label{fig:alignment}
\end{subfigure}
\caption{Compliant motions can be used for aligning both position and orientation of a workpiece}
\label{fig:pos_ori_alignment}
\end{figure}

\fa{Motions that include contact with the environment} can be difficult for robots since pose \fa{(position and orientation)} errors in tasks with small clearance often lead to high contact forces. \change{Two examples of such motions with initial errors are shown in Fig.~\ref{fig:pos_ori_alignment}  (position error in Fig.~\ref{fig:assembly} and orientation error in Fig.~\ref{fig:alignment}).} 
It is essential that the contact wrenches (force and torque) are managed when interacting with the environment; \change{without suitable compliant interaction, the linear motions depicted by the arrows in Fig.~\ref{fig:pos_ori_alignment} would not result in the alignments shown, but instead would cause jamming, wedging, or breakage of equipment or workpieces}. 
Humans, on the other hand, can effectively take advantage of contact forces and utilize the arising compliant motions to mitigate localization uncertainty.
\changetwo{Impedance control is a convenient control approach for compliant motions that does not require switching between different control strategies  \cite{hogan1987stable}}. 
It features a virtual spring with adjustable stiffness between \changetwo{the current and the desired pose}. Impedance control allows small deviations from the desired trajectory, while still applying a stiffness-dependent wrench along the desired trajectory. 
This ability makes impedance controller a natural choice for performing compliant motions. Even though there has been a recent success in developing a feasible trajectory planner for 3-D compliant motions \cite{guan2018efficient}, there is need for end-user friendly learning methods for impedance-controlled compliant motions. \changethree{Interested readers can consult a recent survey \cite{abudakka2020Variable} for different impedance controllers based on variability, control and learning perspectives.}

Learning from Demonstration (LfD) \cite{argall2009survey,osa2018algorithmic} is an established paradigm in robotics for skill transfer and encoding. The key idea is that a human expert gives a demonstration of a task, which the robot then learns to reproduce. There are multiple methods for encoding the learned skill, such as Stable Estimator of Dynamical Systems (SEDS) \cite{khansari2011learning}, Gaussian Mixture Models (GMM) with Gaussian Mixture Regression (GMR) \cite{calinon2007learning}, \changethree{ Riemannian Motion Policies \cite{mukadam2020riemannian}, and several popular movement primitives, such as Dynamic Movement Primitives (DMP) \cite{schaal2006dynamic}, Kernelized Movement Primitives (KMP) \cite{huang2019kernelized} and Probabilistic Movement Primitives (ProMP) \cite{paraschos2013probabilistic}. }
Whereas these methods are perfectly capable of representing free space motions and contact tasks without position uncertainties, \fa{they have a tight coupling between force and position trajectories,}
which makes them susceptible to errors in initial position especially when dealing with multiple demonstrations. 
Even though recent publications have shown that with certain modifications DMPs can be used to realize unseen trajectories \cite{abu2015adaptation}, a primitive without the force-position coupling would be more flexible for easy generalization to tasks similar to demonstration.


\changetwo{This paper extends and generalizes the method presented in \cite{suomalainen2017}. In \cite{suomalainen2017} the goal was to learn a single primitive consisting of a desired direction in translation, and then learn the axes along which compliance was required.} 
\changetwo{
\fa{However, in this paper, we learn} the desired direction, essentially a linear direction in Cartesian space, both for translation and rotation, and the required compliant axes in both translation and rotation as well.
The latter means learning the stiffness matrices for impedance controllers in translation and rotation with certain pre-defined restrictions.} 
The desired direction is defined as a linear \changetwo{6-D} direction which, either through free space or with the help of a mechanical gradient such as a chamfer, leads the end-effector to the goal pose of the motion. 
Kinesthetic teaching is used to show the robot an example of a motion. The key difference to existing LfD methods for in-contact tasks is that the position trajectory is not coupled with the force or impedance profile. This renders the presented primitive more robust against localization errors.

\paragraph{\fa{The novelty in this paper includes:}}
\begin{enumerate}[i)]
  \item \changetwo{extension of \cite{suomalainen2017} to cover also rotational motions and combinations of rotations and translations,}
  \item detecting if either translations or rotations are fully compliant, i.e. no desired direction exists, due to work done by the environment,
  \item evaluating if the desired direction is reliable and finding the compliant axes even when the desired direction is unreliable, and 
  \item \changetwo{showing that, when properly learned with the presented method, the primitive can successfully complete a wide range of linear 6-D motions while taking advantage of the environment as guidance to mitigate localization errors between the tool and the goal.}
\end{enumerate}

Learning the segmenting and sequencing of the primitives to complete a full task is outside of the scope of this paper, but has been shown to be possible in \cite{hagos2018seq}. 

\section{RELATED WORK}
\label{RELATED}
\changetwo{It is a well established idea to use force control for taking advantage of geometry in an assembly task}. \changetwo{The classical work of Mason \cite{mason1981compliance} performed force-controlled peg-in-hole with task frames to avoid the need for the end-user to define low-level commands. } 
Schimmels and Peshkin \cite{schimmels1991force} later defined the concept of geometric force-assemblability in 2-D using screw-theoretical concepts. 
Even earlier, Ohwovoriole and Roth \cite{ohwovoriole1981extension} used the concept of virtual work (defined as the dot product between the motion twist and the contact wrench) to divide twists into repelling, reciprocal or contrary. 
Their research inspired us to look into whether work is done by the human teacher or the environment, which is a key point when discovering whether all translational or rotational degrees of freedom must be compliant. 
\changetwo{For more complicated tasks, it is possible to try detecting the contact formations of the tool \cite{lefebvre2005online}, which allows learning or crafting more elaborate sequences. However, this requires also more information on the tools; thus, the method presented here does not attempt to cover all the problems that can be solved with contact formations, but rather a subset without requiring as much prior information as the use of contact formations.}

In other approaches, Stolt \cite{stolt2015robotic} studied robotic assembly using high-level task specification and alternating position and force control. However, we believe that LfD provides an easier interface for the end-user to teach a task.
\changetwo{One well established idea is to apply reinforcement learning after learning an initial skill with LfD, such as \cite{kalakrishnan2011learning} where exact forces are learned with RL. \fa{However, in this paper, we aim}
to learn the skill without the need for RL; 
the presented method attempts to learn skills that do not require explicit control of contact forces, such as compliant assembly skills, where the method presented in \cite{kalakrishnan2011learning} would be unnecessarily complicated and possibly prone to errors.} 

\change{Most LfD tasks are encoded as motion primitives, such as the earlier mentioned DMP and SEDS, a general way for presenting a trajectory and possibly an additional force profile.
A complex task then consists of a set of primitives, which are triggered in sequence \cite{kroemer2014learning,hagos2018seq}.
The term motion primitive refers then to a model of a phase (a motion segment) of a task, typically modeled by a continuous function approximator aimed towards learning the task from a human demonstration. 
In contrast, this paper proposes a primitive that is specifically targeted to encode phases where contact can assist the task, instead of being a general function approximator.
This allows the proposed primitive to be more robust to localization errors.
}

Learning workpiece alignment from demonstrations using DMP's has been presented by Peternel et al. \cite{peternel2015human}, Deni{\v{s}}a et al.~\cite{denivsa2016learning}, Abu-Dakka et al. \cite{abudakka2020Geometry,abu2015adaptation,abu2018force} and Kramberger et al. \cite{kramberger2016transfer,kramberger2017generalization}. Peternel et al. used an external interface for the teacher to manually modulate the required stiffness. \changethree{Compliant Motion Primitives (CMP) by Deni{\v{s}}a et al.~\cite{denivsa2016learning} and  Abu-Dakka et al. \cite{abu2015adaptation} added a force feedback controller in the DMP's } and lately an impedance profile to GMM's \cite{abu2018force}. Recently, Abu-Dakka  and Kyrki provided geometry-aware DMP's formulation which capable of direct encoding of compliance parameters \cite{abudakka2020Geometry}.
Kramberger et al. performed a peg-in-hole task with varying hole depths, and also performed rotational motions. 
The aforementioned approaches choose different positions regarding a trade-off between accuracy and error tolerance, and have different expectations with regards to localization capabilities of the robot. The DMP-based methods with an impedance profile can achieve an exact level of compliance at a certain point in the trajectory, and perform nonlinear free-space motions; however, the requirement is that the robot's end-effector can be properly localized w.r.t. the environment. If this localization fails due to e.g. camera issues or the goal moving without knowledge of the robot, the impedance profile will be applied at a wrong time.  In contrast, our approach does not attempt an exact control of the impedance profile and relies only on linear motions; whereas this limits the applications, it allows the localization error of the end-effector to match the mechanical convergence region seen in Fig.~\ref{fig:convergence}.


There have been a few other recent publications about new LfD primitives to replace the aforementioned DMP and GMM/GMR strategies. Reiner et al. \cite{reiner2014lat} and Rozo et al. \cite{rozo2013learning} took advantage of the variations in the recorded trajectory to define where pose accuracy is important and therefore high stiffness required. However, their work was aimed towards free space motion and included the whole variance of demonstrations, whereas we look at the variance of motion outside a specified desired direction in an in-contact task. Ahmadzedah et al. \cite{ahmadzedah2017cylinders} proposed an LfD encoding method which can generate unseen trajectories within the cylinder of the given demonstrations. However, both of these methods are presented as tools for free space motion and not for in-contact tasks. Racca et al. \cite{racca2016} used Hidden Semi-Markov Models (HSMM) with GMR to allow the teaching of in-contact tasks. However, even their work cannot take advantage of the task's geometry. The goal of this paper is to present a primitive, and a method for learning the primitive, that can maximize the guidance of a physical gradiennt in the environment.


\section{METHOD}
\label{METHOD}

\begin{figure}[t]
	\centering
	\includegraphics[width=.4 \columnwidth]{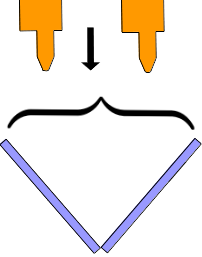}
	\caption{Illustration of the potential convergence region (black brace) of the algorithm in a pure translational case (similar as the "valley" experiment in \cite{suomalainen2017}) \changethree{with the black arrow presenting the direction of motion (the desired direction in translation $\pmb{\hat{v}_{d}^*}$ later in the paper)}. The robot does not need to know where within the black brace the tool is, because the gradients will guide it to the goal. }
	\label{fig:convergence}
\vspace{-1em}
\end{figure}

\changetwo{The method presented here is meant for tasks where localization errors between the tooltip and the goal can grow large; indeed, a main difference to many other LfD methods is that even though other methods can converge to a goal from a wide region, the method presented in this paper does not need to know how far it is from the target, or how far it is from the intended starting pose. For example, in Fig.~\ref{fig:convergence}, the robot does not need to know where within the shown convergence region the end-effector is. The method does not require any feedback, neither force nor pose, during reproduction of a single segment; \changethree{the motion and impedance parameters remain the same during a single segment}. Whereas this greatly simplifies the requirements for the method, there are also disadvantages, such that there is no proper built-in stopping condition for the primitive, raising the need for e.g.~force thresholds to be built into a fully working system and the physical system to have a natural stopping condition. Moreover, the presented primitive can only generate motions that are linear in Cartesian coordinates in free space but might have a different shape if guided by contact.}

\changetwo{
There is no attractor in free space for the primitive. The intuition is to make the robot follow a natural, physical gradient, similarly as a human would intuitively do when faced with localization error. Thus, we assume that even in a demonstration where a human cannot directly align the workpieces but must rely on contact and physical gradients,  there is a \say{correct} direction, called desired direction in this paper; this is the direction where the user would guide the tool if all pose information was correct, that can also be called the working force. This direction is all the information that the robot uses during reproduction -- it does not have explicit knowledge about the initial pose or the goal pose. 
Additionally, there is no in-built upper limit for the forces, which will grow in accordance with the stiffness. For example, if the contact is made near the edge of the convergence region depicted in Fig.~\ref{fig:convergence}, the force exerted at the bottom of the valley will be higher than if contact is made closer to the bottom. However,  as the goal of this primitive is alignment, this is an acceptable side-effect, and dangerously large forces could be avoided with a simple threshold.  }

It is assumed that an assembly task can be divided into motion segments which can be completed with combinations of linear motion and compliance; \changetwo{in previous work \cite{hagos2018seq} it was shown that 1) a demonstration for tasks such as hose coupling can be automatically divided into segments which the presented primitive can complete and 2) the contact transitions can be automatically learned for reproducing the motions.}

\change{The controller parameters are learned offline after a demonstration}.
To complete the task, each segment can be executed with an impedance controller defined for the end-effector as
\begin{equation}
  \begin{split}
  & \pmb{F} = K_f(\pmb{x}^*-\pmb{x})-D_f\pmb{v} \\
  & \pmb{T} = K_o(\log(B^T B^*))-D_o\pmb{\omega}
  \end{split}
  \label{eqt:imp_control}
\end{equation}  
where $\pmb{F,T}$ are the force and torque used to control the robot, $\pmb{x}^*$ the desired position, $\pmb{x}$ the current position, \change{$B^*,B\in SO(3)$ rotation matrices representing the desired orientation and the current orientation, and $\log(\cdot)$ denotes the \fa{rotation} matrix logarithm}. $K_f$ and $K_o$ are stiffness matrices and $D_f\pmb{v}$ and $D_o\pmb{\omega}$ damping terms. \change{Notation summary can be found from Table~\ref{tab:notation}.}

\begin{table}[t]
  \centering
  \begin{tabular}{@{}cc @{}}
  
  	{\bf \change{Symbol}} & {\bf \change{Meaning}} \\ \cmidrule{1-2}    
  		{$\pmb{F}$} & {\change{Commanded Cartesian force}} \\ \cmidrule{1-2}
  		{$\pmb{F}_{ext}$} & {\change{Force applied by the teacher}} \\ \cmidrule{1-2}
  		{$\pmb{F_m}$} & {\change{Measured force}} \\ \cmidrule{1-2}
  		{$K_f$} & {\change{Translational stiffness matrix}} \\ \cmidrule{1-2}
        {$K_o$} & {\change{Rotational stiffness matrix}} \\ \cmidrule{1-2}
        {$s$} & {\change{Sector of desired directions}} \\ \cmidrule{1-2}
        {$\pmb{T}$} & {\change{Commanded torque}} \\ \cmidrule{1-2}
        {$U_d$} & {\change{Rank $d$ PCA approximation}} \\ \cmidrule{1-2}
        {$\pmb{v}$} & {\change{Translational velocity}} \\ \cmidrule{1-2} 
        {$\pmb{v}_a$} & {\change{Actual direction of motion in translation}} \\ \cmidrule{1-2}
        {$\lambda$} & {\change{Rotational speed}} \\ \cmidrule{1-2}
        {$\nu$} & {\change{Translational speed}} \\ \cmidrule{1-2}
        {$\pmb{\Pi}$} & {\change{Measured force or torque}} \\ \cmidrule{1-2}
        {$\pmb{\overline{\psi}_{a}}$} & {\change{Actual direction of motion (translation or rotation)} } \\ \cmidrule{1-2}
        {$\pmb{\hat{\omega}_{d}^*}$} & {\change{Desired direction in rotation}} \\ \cmidrule{1-2}
        {$\pmb{\hat{v}_{d}^*}$} & {\change{Desired direction in translation}} \\ \cmidrule{1-2}
        {$D$} & {\change{Damping}} \\ \cmidrule{1-2} 
        {$\pmb{x}$} & {\change{Position}} \\ \cmidrule{1-2}
        {$B$} & {\change{A rotation matrix}} \\ \cmidrule{1-2}

  \end{tabular}
  \caption{\change{A notation summary of frequently used symbols.}}
  \label{tab:notation}
\end{table}

As each segment consists of an impedance controller primitive, we calculate the desired trajectory for each segment in a feed-forward manner 
\begin{equation}
\label{eqt:feedforward}
  \begin{split}
  & \pmb{x}^*_t= \pmb{x}^*_{t-1}+\nu  \Delta t\pmb{\hat{v}_{d}^*}   \\
  & B^*_t=B^*_{t-1}\exp( \lambda  \Delta t \left[ \pmb{\hat{\omega}_{d}^*} \right] ) 
  \end{split}
\end{equation} 
where $\pmb{\hat{v}_{d}^*}$ and $\pmb{\hat{\omega}_{d}^*}$ are the desired directions in translation and rotation, $[\cdot]$ denotes the skew symmetric matrix corresponding to a vector,  $\Delta t$ the sample time of the control loop and $\nu$ and $\lambda$ the translational and rotational speeds. Throughout this paper, we will use the circumflex (\^{}) notation to denote the normalization of a vector (i.e. $\pmb{\hat{x}}=\frac{\pmb{x}}{|\pmb{x}|}$) \changetwo{, the subscript $_d$ when referring to \say{desired} direction and $_a$ to \say{actual}, the latter meaning the observed direction of motion, either translation of rotation.}

\fa{In this paper,} we propose a method to learn $K_f$, $K_o$, $\pmb{\hat{v}_{d}^*}$ and $\pmb{\hat{\omega}_{d}^*}$ separately for each primitive from one or more human demonstrations, assuming that the damping is sufficient for stabilizing the dynamics, for example by manually tuning the damping parameters $D_f$ and $D_o$ to avoid instabilities.  \changetwo{$K_f$ and $K_o$ will be learned with the restrictions that stiffness of an axis is either \say{stiff} with an application-dependent stiffness value $k$, or 0, rendering this axis compliant. Thus, these parameters will lead the tool into a certain direction, accompanied with suitable stiffness parameters, that can take advantage of physical guides to lead the tool into a final pose. We note that the algorithm does not explicitly know this final pose, but there needs to be a physical limitation that eventually stops the end-effector, or a separate thresholding method to detect success.}

In cases such as depicted in Figs.~\ref{fig:assembly} or \ref{fig:alignment}, giving at least one demonstration from each shown starting position allows the algorithm to learn a set of parameters which can reproduce all motions from within the workpiece's zone of convergence. \change{The zone of convergence can be considered as the set of workpiece poses from which a mechanical gradient can lead the workpiece to the goal position; a simple case is visualized in Fig.~\ref{fig:convergence}. Without correct compliance, the tool will get stuck or misaligned upon reaching contact. Thus, the use of interaction forces as guidance requires task-specific compliance.} 

\begin{figure}[t]
	\centering
	\includegraphics[width=.8\columnwidth]{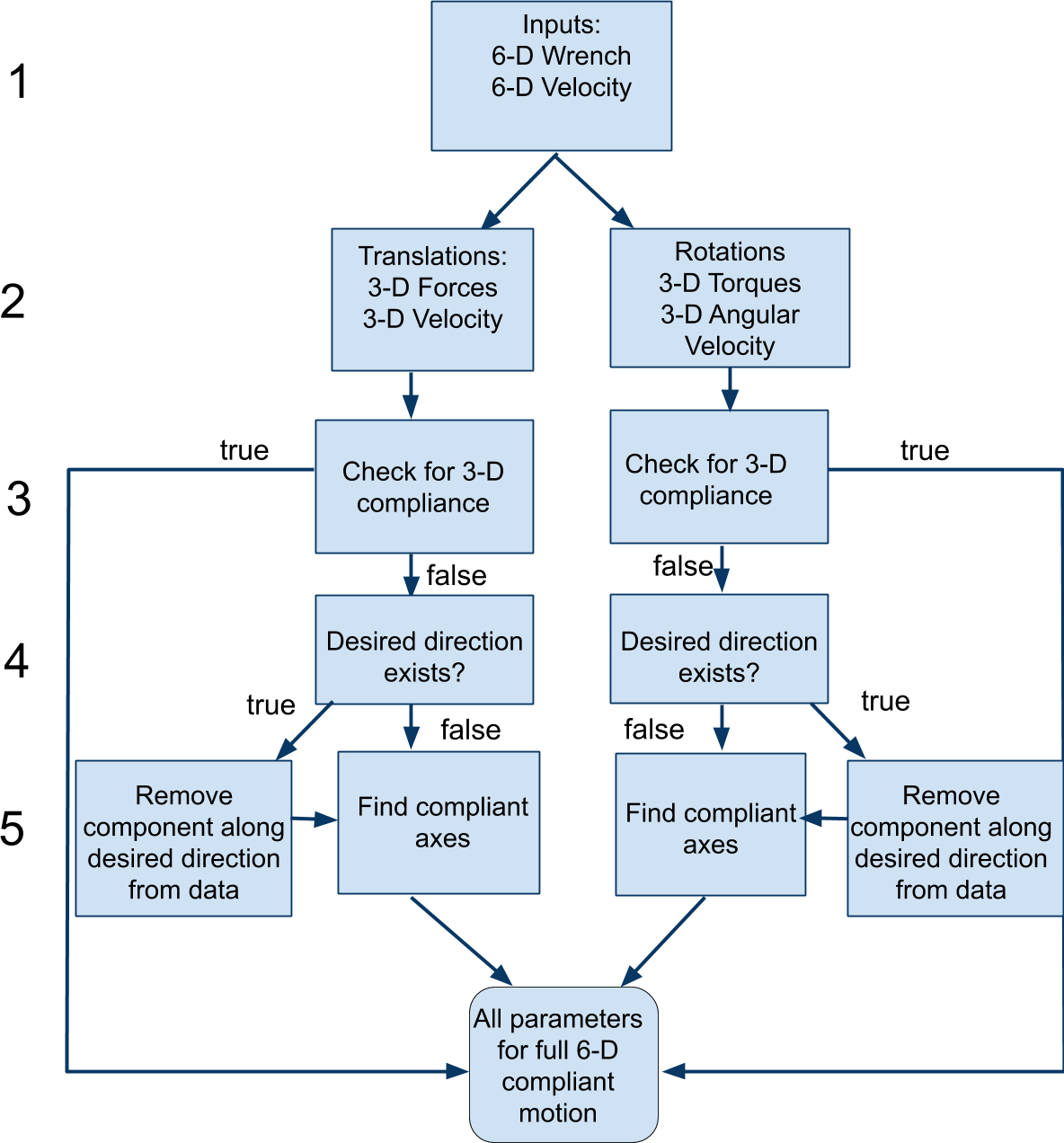}
	\caption{A flowchart describing the whole process of finding the 6-D compliant primitive to reproduce a demonstrated motion. The numbers on the left hand side present different stages of the algorithm for clarification.} 
	\label{fig:flow}
\vspace{-1em}
\end{figure}

\change{A flowchart describing the whole process of learning  $K_f$, $K_o$, $\pmb{\hat{v}_{d}^*}$ and $\pmb{\hat{\omega}_{d}^*}$ from a demonstrated motion is shown in Fig.~\ref{fig:flow} with numbers presenting stages of the algorithm. In Section \ref{sec:full_compl}, matching stage 3 of the flowchart, we validate whether the teacher performed only translation, i.e. $\pmb{\hat{\omega}_{d}^*}$ is zero (does not exist) even though rotation was observed. This results in all rotational degrees of freedom to be compliant, or vice versa if the teacher performs rotation only (called 3-DOF compliance in this paper). 
If this is not observed, in Section \ref{sec:desired} \fa{which corresponds to stage 4 of the flowchart,}
the algorithm computes $\pmb{\hat{v}_{d}^*}$ and $\pmb{\hat{\omega}_{d}^*}$, or validates if either of them is not required. 
Finally, in Section \ref{sec:compliance} matching stage 5 of the flowchart, it is evaluated if individual degrees of freedom are required to be compliant, yielding $K_f$ and $K_o$. As an end result, there can be a desired direction in both translation and rotation, or in only one of them. In addition, compliance is found for both rotation and translation, if required.}

\begin{figure}[t]
	\centering
	\includegraphics[width=.8\columnwidth]{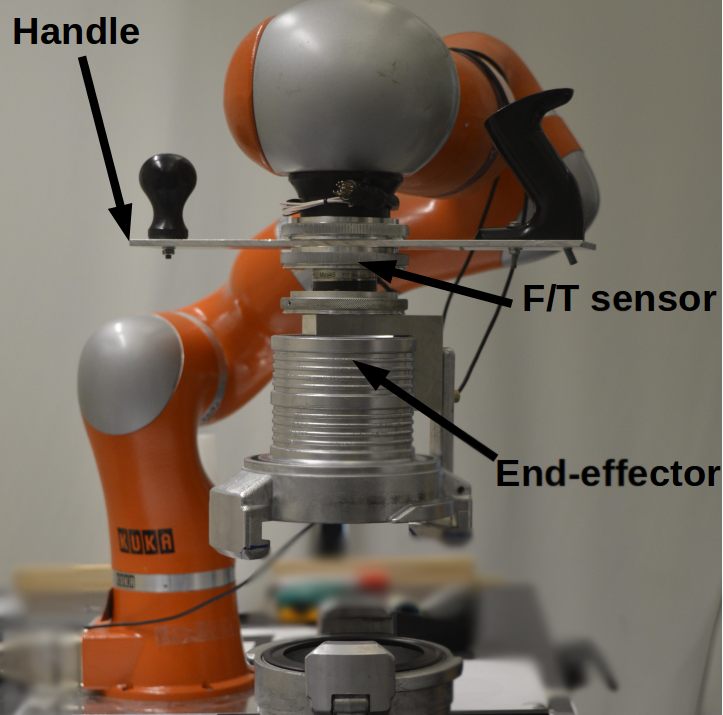}
	\caption{ The KUKA LWR4+ robot used for the experiments, with equipment for the hose-coupler setup attached.}
	\label{fig:hose_coupler}
\vspace{-1em}
\end{figure}

The method requires that during the demonstration, a force/torque (F/T) sensor is placed between the tool and the place where the teacher grabs the robot, such as in Fig.~\ref{fig:hose_coupler}. Wrench and pose data at the F/T sensor are recorded, and the force measured by the F/T sensor during contact (neglecting Coriolis and centrifugal force) can be written as 
\begin{equation}
  \pmb{F_m} = \pmb{F_N} + \pmb{F_{\mu}} + m\pmb{a}
  \label{eqt:measured_force}
\end{equation}
where $\pmb{F_m}$ is the force measured by the F/T sensor, $\pmb{F_N}$ the normal force, $\pmb{F_{\mu}}=\vert\mu\pmb{F_N}\rvert\left( -{\pmb{\hat{v}_a}}\right)$ the force caused by Coulomb friction with $\mu$ being the friction coefficient and $\pmb{\hat{v}_a}$ the actual direction of motion, $m$ the mass of the tool and $\pmb{a}$ it's acceleration. Similarly, the measured torque $\pmb{T_m}$ can be written as
\begin{equation}
  \pmb{T_m} = \pmb{\rho} \times \pmb{F_N} + \pmb{l} \times \pmb{F_{\mu}} + I\pmb{\alpha}
  \label{eqt:measured_torque}
\end{equation}
where $\pmb{l}$ and $\pmb{\rho}$ are the lever arm position vectors perpendicular to corresponding applied forces, $I$ the inertia matrix and $\pmb{\alpha}$ the angular acceleration. Although this model is for a single-point contact, we show that the method is robust enough that we can teach multi-point contact tasks as well; considering a thorough contact formation treatment is outside the scope of this paper. We assume that the speed of the end-effector is close to constant and therefore the acceleration terms can be ignored from both equations.  

\subsection{Checking for 3-DOF compliance}
\label{sec:full_compl}
In 6-D motion it is possible that, due to contact forces, translational force applied by the teacher causes rotation, or vice versa. In such a case, either the observed translation or rotation is caused completely by the environment and the corresponding degrees of freedom need to be set compliant (i.e. 3-DOF compliance). More insight into the kind of motion falling into this category can be found from Fig.~\ref{fig:reuna} and Section \ref{sec:dof}.

The intuition to detect this phenomenon stems from the definition of work in physics, which is defined for translational and rotational motions as
\begin{equation}
\begin{split}
W_{x} &= \pmb{F_m} \cdot \Delta \pmb{x} \\
W_{\beta} &= \pmb{T_m} \cdot \Delta\pmb{\beta}
\end{split}
\end{equation}
where $W$ is the work, $\Delta \pmb{x}$ the change in translation and $\Delta\pmb{\beta}$ the change in angle. If the majority of work is done by the environment, we assume that those degrees of freedom (all rotational or translational degrees of freedom) should be compliant since the demonstrator was not explicitly performing those motions but they were caused by the environment. Formally, either rotation or translation is 3-DOF compliant if
\begin{equation}
\frac{W_{env}}{W_{tot}} \geq \sigma
\label{eqt:full_compl}
\end{equation}
where $W_{tot}$ is the total work during a demonstration and $W_{env}$ the work done by the environment. We can compute $W_{tot}$ by
\begin{equation}
W_{tot} = \int \lvert W \rvert dt
\end{equation}
where $W$ is either $W_{x}$ or $W_{\beta}$ and taking the absolute value means that we consider work to be path-dependent. As the wrench measured by the F/T sensor is the contact wrench, i.e. caused by the environment, work performed by the environment is observed as positive values for $W$. Therefore we can compute $W_{env}$ as
\begin{equation}
W_{env} = \int W_+dt, \ W_+ =
\begin{cases}
W \ \mathrm{if} \ W>0 \\
0 \ \mathrm{else}
\end{cases}
\end{equation}

The choice of $\sigma$ in (\ref{eqt:full_compl}) depends on the task and the accuracy of demonstration; with perfect demonstrations $\sigma$ could be set to 1, but in practice it has to be reduced to allow human inconsistencies during a demonstration. If the ratio is below $\sigma$, Algorithm \ref{alg:vd} is run as described in Fig.~\ref{fig:flow} and in the next section. Otherwise rotation or translation is set to 3-DOF compliant.

\subsection{Learning desired direction}
\label{sec:desired}

\begin{figure}[t]
	\centering
	\includegraphics[width=.9\columnwidth]{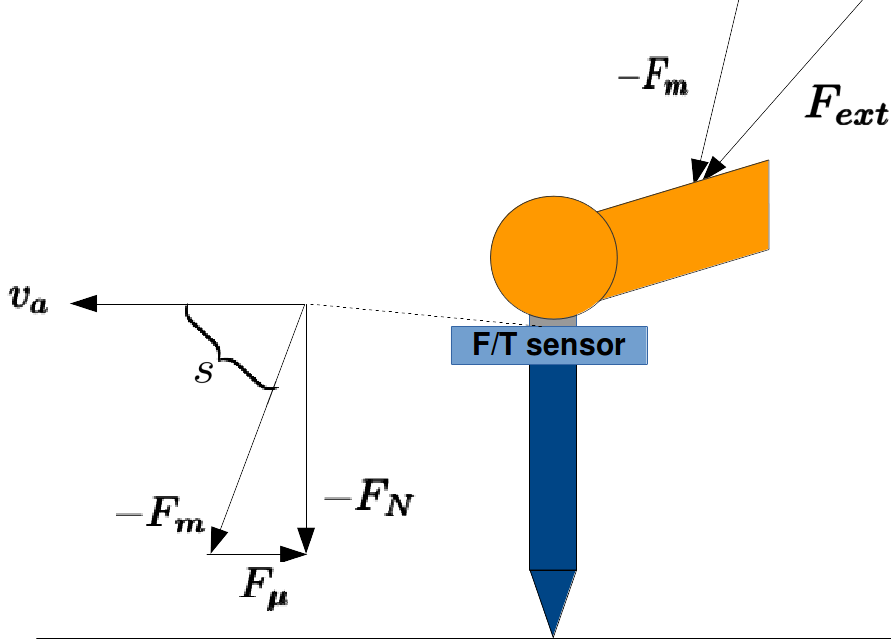}
	\caption{Force/torque sensor configuration, the position where \changetwo{the working force} by the human teacher $\pmb{F_{ext}}$ is applied and the forces which sum up to the reading of the force measurement $\pmb{F_{m}}$ of the F/T sensor.}
	\label{fig:sliding_forces}
\vspace{-1em}
\end{figure}

In this section we describe the method to learn $\pmb{\hat{v}_{d}^*}$ and $\pmb{\hat{\omega}_{d}^*}$. To slide the robot's tool in contact, the robot can be pushed from any direction from the sector $s$ defined as the 2-D sector between the actual direction of motion $\pmb{v_a}$ and the force measured by the F/T sensor $\pmb{F_m}$, as seen in (\ref{eqt:measured_force}) and Fig.~\ref{fig:sliding_forces}. 
\changetwo{Thus, different directions and magnitudes of force may result in the same observed trajectory, which is sliding along the surface: this means that when contact is leveraged properly, different actual motions can be realized even when the robot is applying the same force. The key idea is to use this insight to narrow down the possibility of where the teacher's working force is applied, such that the same working force can cause different motion directions if required}.
This idea is extended into rotations and 3-D such that at each measurement point of a demonstration, we find a set of force and torque directions which would result in the observed direction of motion. 
\change{Issues such as curved surfaces and possible surface artifacts, however, often cause the set to vary even along a single motion.} By taking an intersection over many such sets, \changetwo{we can find a direction that could have created the direction at any point during the demonstration} and thus reproduce motions which can be represented with linear impedance controller parameters. The same algorithm, presented in Algorithm~\ref{alg:vd} and explained in the upcoming paragraphs, is used to find both $\pmb{\hat{v}_{d}^*}$ and $\pmb{\hat{\omega}_{d}^*}$. 

\begin{figure}[t]
\centering
\includegraphics[width=.6\columnwidth]{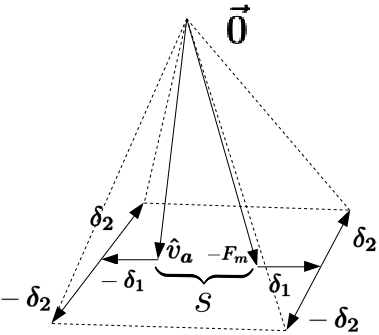}
\caption{Illustration of expanding 2-D sector $s$ for translations into 3-D set of directions $P$ in (\ref{eqt:epsilons}) and (\ref{eqt:polyhedron}). Continuous lines represent the vectors and dotted lines highlight the pyramid shape. }\label{fig:single_pyramid}
\end{figure}

To find an intersection of sectors over a real demonstration in 3-D, sector $s$ must be expanded since a human cannot perform a perfect demonstration (for example, sliding along a straight line on a surface). We expand the sector both perpendicular to $s$ and along the direction of $s$, as seen in Fig.~\ref{fig:single_pyramid}. 

Formally, we define the vectors extending the sector $s$ at each time step $t$ as 
\begin{equation}
\begin{split}
\pmb{\delta}_{1,t} = \tan \xi \frac{-\pmb{\widehat{\Pi}_t} - \pmb{\widehat{\psi}_{a,t}} }{\lvert -\pmb{\widehat{\Pi}_t} - \pmb{\widehat{\psi}_{a,t}} \rvert } \\
\pmb{\delta}_{2,t} = \tan \eta \frac{-\pmb{\widehat{\Pi}_t} \times \pmb{\widehat{\psi}_{a,t}} }{\lvert -\pmb{\widehat{\Pi}_t} \times \pmb{\widehat{\psi}_{a,t}} \rvert } 
\end{split}
\label{eqt:epsilons}
\end{equation}
where \change{$\pmb{\widehat{\Pi}}$ represents either force or torque and $\pmb{\widehat{\psi}}$ either translational or rotational motions, such that the equation is either only translations or only rotations, i.e.~}$\left(\pmb{\widehat{\Pi}}, \pmb{\widehat{\psi}}\right) \in \lbrace \left(\pmb{\widehat{F}_m}, \pmb{\hat{v}} \right) , \left(\pmb{\widehat{T}_m}, \pmb{\hat{\omega}} \right) \rbrace$.  Variable $\eta$ is the angle with which we wish to extend the sector $s$ perpendicularly and $\xi$ the angle used to widen the sector. Thus the limits of a desired direction of motion, as illustrated in Fig.~\ref{fig:single_pyramid} for translations, at each time step $t$ can be written as a set of vectors
\begin{equation}
\begin{aligned}
P_t= 
\{
\pmb{\widehat{\psi}_{a,t}}-\pmb{\delta_{1,t}}+\pmb{\delta_{2,t}},\,
\pmb{\widehat{\psi}_{a,t}}-\pmb{\delta_{1,t}}-\pmb{\delta_{2,t}},\, \\
\pmb{-\widehat{\Pi}_t}+\pmb{\delta_{1,t}}+\pmb{\delta_{2,t}},\
\pmb{-\widehat{\Pi}_t}+\pmb{\delta_{1,t}}-\pmb{\delta_{2,t}}
 \}
\label{eqt:polyhedron}
\end{aligned}
\end{equation}
where $P_t$ represents the set of vectors limiting the desired directions of motion $\pmb{\widehat{\psi}_{d,t}}$ at a single time step $t$ in 3-D, either translational or rotational. Thus, we can write the range of possible desired directions at time step $t$ as a positive linear combination of the vectors in $P_t$. In Algorithm \ref{alg:vd} the computation of each $P_t$ is shown on line \ref{line:pt}.

\begin{algorithm}[t]
	\caption{Computation of desired direction}
	\hspace*{\algorithmicindent} \textbf{Input:} Sets $\Psi,\Pi$ consisting of $\pmb{\widehat{\psi}_{a,t}},\pmb{\widehat{\Pi}_t}$. \\
	\hspace*{\algorithmicindent} \textbf{Output:} Desired direction $\pmb{\widehat{\psi}_{d}^*}$.
	\label{alg:vd}
	\begin{algorithmic}[1]
		\State{Determine $R$ that rotates mean $\pmb{\overline{\psi}_{a}}$ of all $\pmb{\widehat{\psi}_{a,t}}$ to positive z axis} \label{line:buildr}
		\tikzmk{A}
			\For {each measurement point $t$} \Comment{3-D $\rightarrow$ 2-D}
			\State{Calculate $P_t$ from $\pmb{\widehat{\psi}_{a,t}},\pmb{\widehat{\Pi}_t}$ (\ref{eqt:epsilons}),(\ref{eqt:polyhedron})} \label{line:pt} 
			\For {each $\pmb{p_i}$ in $P_t$}
			\State{$\Theta_t=\Theta_t \cup \mathbf{vec2ang}(R\pmb{p_i})$} \label{line:rotate}
			\EndFor
			\State{$\Theta=\Theta \cup \Theta_t $ } 
			\EndFor
		\tikzmk{B}\boxita{green1}
		\State{$G(i,j)=0$ $\forall\; i,j$} \label{line:startpointsearch}
		\tikzmk{B}
			\For {each $\Theta_t \in \Theta$} \Comment{Vote map for outlier rejection}
			\For {$(i,j)$ inside rectangle formed by $\Theta_t$}
			\State{$G(i,j) = G(i,j) + 1$}
			\EndFor
			\EndFor 
		\tikzmk{C}\boxitb{pink1}
		\State{$\displaystyle{( i,j )_{max} = \arg\max_{g} \: \forall\, g_{i,j}\ \mathrm{in}\ G}$} \label{line:endpointsearch}
		\tikzmk{C}
			\For {each $\Theta_t$ in $\Theta$} \label{line:startchoosing} \Comment{Choose only inliers}
			\If{$(i,j)_{max}$ inside $\Theta_t$}
			\State{$\Theta^*=\Theta^*\cup \Theta_t$}
			\EndIf
			\EndFor 
		\tikzmk{D}\boxitc{blue1}
		\State{$\displaystyle{\Phi = \bigcap_t \Theta^*_t}$} \label{line:endchoosing} \Comment{Compute intersection}
		\State{Calculate Chebyshev center $\pmb{\phi^*}$ of $\Phi$} 
		\State{$\pmb{\widehat{\psi}_{d}^*} = R^{-1}\mathbf{ang2vec}(\pmb{\phi^*})$} \Comment{2-D $\rightarrow$ 3-D}
	\end{algorithmic}
\end{algorithm}

To avoid problems due to representation of orientation, the data is rotated on lines \ref{line:buildr} and \ref{line:rotate} in Algorithm \ref{alg:vd}. \change{As taking an intersection over 3-D polyhedra is computationally expensive, each 3-D data point is projected into 2-D unit circle using function \textbf{vec2ang} described in Algorithm \ref{alg:vec2ang}. Essentially, this step projects the 3-D pyramids $P_t$, shown in Fig.~\ref{fig:single_pyramid}, into 2-D rectangles $\Theta_t$ (more details can be found in \cite{suomalainen2017}).}   

On lines \ref{line:startpointsearch}-\ref{line:endpointsearch} in Algorithm \ref{alg:vd} outlier rejection is performed: we find, on a chosen scale, the point $(i,j)$ of grid $G$ which is enclosed by the maximum number of rectangles $\Theta_t$. Then on lines \ref{line:startchoosing}-\ref{line:endchoosing}  we choose from the set of rectangles $\Theta$ the subset $\Theta^*$ which include the point $(i,j)$. Then we compute the intersection $\Phi$ of rectangles $\Theta^*$, compute the Chebyshev center \cite{garkavi1964chebyshev} $\pmb{\phi^*}$ of $\Phi$, convert $\pmb{\phi^*}$ back to a 3-D vector with  function \textbf{ang2vec} (Algorithm \ref{alg:ang2vec}) and rotate it back to get $\pmb{\widehat{\psi}_d^*}$. The process is similar to \cite{suomalainen2017}, where it is explained in more detail. 

\begin{algorithm}[t]
\caption{\textbf{vec2ang()}.}
\hspace*{\algorithmicindent} \textbf{Input:} Cartesian 3-D vector $\pmb{p}$. \\
\hspace*{\algorithmicindent} \textbf{Output:} Angular 2-D vector $\pmb{\theta}$.
\label{alg:vec2ang}
\begin{algorithmic}[1]
\State {$\pmb{\hat{p}} = \frac{\pmb{p}}{\lvert \pmb{p} \rvert}$}  
\State $r = \arccos(\hat{p}_z)$ \\
$\gamma = \arctan2(\hat{p}_y,\hat{p}_x)$ \\
$\theta_x = r\cos(\gamma)$ \\
$\theta_y = r\sin(\gamma)$
\end{algorithmic}
\end{algorithm}

\begin{algorithm}[t]
\caption{\textbf{ang2vec()}.}
\hspace*{\algorithmicindent} \textbf{Input:} Angular 2-D vector $\pmb{\theta}$. \\
\hspace*{\algorithmicindent} \textbf{Output:} Cartesian 3-D vector $\pmb{p}$.
\label{alg:ang2vec}
\begin{algorithmic}[1]
\State{$S = \textrm{sign}(\arctan2(\theta_y,\theta_x))$} \\
$r = \cos\left(\sqrt{\theta_x^2+\theta_y^2}\right)$ \\
$a = \frac{\theta_y}{\theta_x}$ \\
$\hat{p}_z = \arccos(r)$ \\
$\hat{p}_x = S\sqrt{\frac{1-\hat{p}_z^2}{1+a^2}}$ \\
$\hat{p}_y = S a \hat{p}_x$
\end{algorithmic}
\end{algorithm}
Since a motion can consist of both translation and rotation, it is possible that for either translation or rotation there does not exist a desired direction, even if 3-DOF compliance is not detected in (\ref{eqt:full_compl}). This can be evaluated from the ratio of outliers i.e. the ratio between the number of rectangles in the set that contributed to the computation of $\Phi$, $\Theta^*$, and all the rectangles $\Theta$. If this ratio is low, it means that there has been a large number of outliers and therefore the corresponding $\pmb{\widehat{\psi}_{d}^*}$ is unreliable. Formally, we assume there is no desired direction if
 
\begin{equation}
 \frac{\lvert\Theta^*\rvert}{\lvert\Theta\rvert} \leq \zeta
\label{eqt:prop_vd}
\end{equation}
where $\zeta$ is a threshold for the ratio and $\lvert\cdot \rvert $ denotes the cardinality of a set, i.e. the number of elements in it. With perfect demonstrations $\zeta$ can be set to 1. However, due to measurement errors, noise and imperfect demonstration, a choice must be made depending mainly on the number of demonstrations and the environment. The key point in choosing $\zeta$ is that higher values demand higher precision from the demonstrations and may discard a detected desired directions, whereas lower values may cause false positives. If, for example, two demonstrations are given from opposite sides such as in Fig.~\ref{fig:assembly}, the value of $\zeta$ should be over 0.5 to ensure that there exists a common desired direction for the two demonstrations. However, in an environment with high friction the threshold may have to be lowered since high friction reduces the width of sector $s$ from Fig.~\ref{fig:sliding_forces}. If the ratio for either translations or rotations is below $\zeta$, then there is no motion in those degrees of freedom. Whether compliance is required along particular axes is tested as described in the next section, and the non-compliant axes will be set stiff; \changetwo{the exact value for this stiffness depends on the application, but in essence this axis is at least close to pure position control.}

Finally, if both $\pmb{\hat{v}_d^*}$ and $\pmb{\hat{\omega}_d^*}$ exist, the ratio between rotational and translational motion must be calculated from unnormalized data. Borrowing from screw theory, we call this value the pitch, defined as

\begin{equation}
\pi=\frac{d_{x}}{d_{\beta}}
\label{eqt:pitch}
\end{equation}
where $\pi$ is the pitch, $d_{x}$ is the \changetwo{ translational distance covered during the motion and $d_{\beta}$ the amount of degrees rotated during a motion used for learning one primitive}. \changetwo{$\nu$ and $\lambda$ can be used to modify the execution speed of the robot as the user wants, but they must be set such that $\nu=\pi\lambda$.} We want to note the possibility that $\pmb{\widehat{\psi}_{d}^*}$ is found in a case where the task requires keeping either rotations or translations only stiff. In such a case the pitch $\pi$ is important: it will make the velocity small enough that the motion in reproduction is minimal, essentially keeping those degrees of freedom stiff.

\subsection{Learning axes of compliance}
\label{sec:compliance}
This section presents how to learn $K_f$ and $K_o$ such that, together with $\pmb{\hat{v}_d^*}$ and $\pmb{\hat{\omega}_d^*}$, the demonstrated motion can be reproduced. Our key assumption for detecting the axes of compliance is that if there is motion in other directions besides $\pmb{\hat{\psi}_{d}^*}$, that motion must be caused by the environment, signalling a direction where compliance is required. We assume that if compliance is required along an axis, it must be totally compliant (i.e. stiffness equals zero). Hence if $\pmb{\hat{v}_d^*}$ exists, the axes of compliance defined in $K_f$ must be perpendicular to $\pmb{\hat{v}_d^*}$, and similarly for $\pmb{\hat{\omega}_d^*}$ and $K_o$; \change{the robot arm would not move towards a direction with zero stiffness, even if commanded to}. We find the directions of the compliant axes with the help of Principal Componen Analysis (PCA). We compute likelihoods of how well each PCA vector fits the data and based on that decide which of the PCA vectors need to be compliant. The whole process for defining the compliant axes is presented in Algorithm \ref{alg:compliant}.

\begin{algorithm}
\caption{Finding the required number of compliant axes and their directions.}
\hspace*{\algorithmicindent} \textbf{Input:} Desired direction $\pmb{\widehat{\psi}_{d}^*}$, matrix $\overline{\Psi}_a$ consisting of mean directions $\pmb{\overline{\psi}_{a,j}}$ from each $j$ demonstration. \\
\hspace*{\algorithmicindent} \textbf{Output:} $D$ number of compliant axes and $U_D$ their directions.
\label{alg:compliant}
\begin{algorithmic}[1]
\If {$ \pmb{\widehat{\psi}_{d}^*} \neq \varnothing $} 
\For{ $\pmb{\overline{\psi}_{a,j}} \in \overline{\Psi}_a$}
\State {$\pmb{\overline{\psi}_{a,j}}= \pmb{\overline{\psi}_{a,j}} - \left( \pmb{\overline{\psi}_{a,j}} \cdot \pmb{\widehat{\psi}_{d}^*}\right) \pmb{\widehat{\psi}_{d}^*}$} \label{line:removedd}
\EndFor
\EndIf
\For{$d=0\dots 3$ degrees of freedom} \label{line:likelstart}
\State{$U_d=$ rank $d$ PCA approximation of $\overline{\Psi}_a$}
\For{$\pmb{\overline{\psi}_{a,j}}$ in $\overline{\Psi}_a$}
\State{$\pmb{\epsilon_{d,j}} = \left( I-U_d \right)  \pmb{\overline{\psi}_{a,j}}$}
\EndFor
\State{$L_d = \prod \limits_{j} \mathcal{N}\left(\pmb{\epsilon_{d,j}}\vert \pmb{0},\Sigma\right)$} \label{line:likelend}
\State{Calculate $BIC_d$ with $L_d$,(\ref{eqt:bic})} \label{line:bicstart}
\EndFor
\State{$D = \arg \min_d BIC_d$} \label{line:bicstop}

\end{algorithmic}
\end{algorithm}

To enforce the orthogonality between $\pmb{\widehat{\psi}_{d}^*}$ and the axes of compliance when~$\pmb{\widehat{\psi}_{d}^*}$ exists, we remove the component along $\pmb{\widehat{\psi}_{d}^*}$ from the mean of actual motion~$\pmb{\overline{\psi}_a}$ by the computation on line 3 in Algorithm \ref{alg:compliant}. Now any non-zero values of $\pmb{\overline{\psi}_a}$ correspond to motion outside the direction of $\pmb{\widehat{\psi}_{d}^*}$.

Our idea is to validate how many degrees of freedom are required to explain $\pmb{\overline{\psi}_a}$ by calculating the likelihoods $L_d$ for each $d$ number of compliant axes. These degrees of freedom can be understood roughly as the number of linear directions of motion caused by the environment. We use PCA to find the eigenvectors i.e. directions of maximum variance of the data such that they form an orthonormal base. If $\pmb{\overline{\psi}_a} \approx \pmb{0}$, then $\pmb{\overline{\psi}_a}$ is best explained by the origin only, corresponding to $U_d=U_0$ (i.e.\ a rank 0 matrix, meaning zero matrix) and meaning that no compliance is required. If one axis of compliance is required, all motion $\pmb{\overline{\psi}_a}$ has been along a single line, the first principle component corresponding to rank 1 PCA approximation $U_1$. For two axes of compliance, the plane described by the first two principal components best explains the motions. Finally, if not even a plane can explain the data, we require all three axes to be compliant, which can only happen if there is no $\pmb{\widehat{\psi}_{d}^*}$. These computations happen on rows \ref{line:likelstart}-\ref{line:likelend} on Algorithm \ref{alg:compliant}. 

Since we wish to give preference to simpler models, for choosing the final $D$ we take inspiration from Bayesian Information Criterion (BIC) \cite{schwarz1978estimating}, which is defined
\begin{equation}
BIC = \ln(n)k-2\ln(L)
\label{eqt:bic}
\end{equation}
where $n$ is the number of data points, $k$ the number of parameters and $L$ the likelihood of a model. Now we can choose the correct model on rows \ref{line:bicstart}-\ref{line:bicstop} on Algorithm \ref{alg:compliant}. 

It should be noted that the proposed approach does not follow the typical use of BIC which is only applicable when $n\gg k$ and the variance in the likelihood is calculated from the data. Instead, we assume that the uncertainty of demonstrations can be estimated beforehand, making it possible to use the proposed formulation. Also, we note that although here the three axes of compliance outcome is the same as from (\ref{eqt:full_compl}) in Section \ref{sec:full_compl}, the mechanism behind these outcomes is different: without calculating (\ref{eqt:full_compl}) in Section \ref{sec:full_compl}, the method in Section \ref{sec:desired} can detect a desired direction for translations in a case where the cause is actually rotation and the normal force of the environment, or vice versa. Therefore, these two methods are not overlapping.  

If more than one demonstrations are given, the demonstrations are concatenated and the method works exactly the same way. The number of required demonstrations depends on the application and the quality of the demonstrations: with good demonstrations, no more than one demonstration from each approach direction is required. However, there is a lower bound: Algorithm \ref{alg:compliant} cannot detect more degrees of freedom than provided demonstrations. Therefore to take advantage of geometrical properties of the task such as in Fig.~\ref{fig:assembly} and \ref{fig:alignment}, at least two demonstrations are required. It should also be noted that with only one demonstration, (\ref{eqt:bic}) is not applicable since the first term will always go to zero.

\section{EXPERIMENTS AND RESULTS}
\label{EXPERIMENTS}

We used a KUKA LWR4+ lightweight arm to test our method. The demonstrations were recorded in gravity compensation mode, where the robot's internal sensors recorded the pose of the robot and an ATI mini45 F/T sensor placed at the wrist of the robot recorded the wrench. We implemented our controller through the Fast Research Interface (FRI) \cite{schreiber10}, where the controller can be executed as

\begin{equation}
    \begin{split}
  \pmb{\tau}=J^T & (\mathrm{diag}(\pmb{k_{FRI}})(\pmb{x^*}-\pmb{x})\\
            &+\mathrm{diag}(\pmb{d_{FRI}})\pmb{v}+\pmb{F_{FRI}})+\pmb{f_{dyn}}
    \end{split}
  \label{eqt:kuka}
\end{equation}
where $J$ is the Jacobian, $\mathrm{diag}(\pmb{k_{FRI}})$ a diagonal matrix constructed of the gain values of $\pmb{k_{FRI}}$, $\pmb{x^*}-\pmb{x}$ the difference between commanded and actual position and $\pmb{\tau}$ the commanded joint torques. We implemented our controller through the superposed Cartesian wrench term $\pmb{F_{FRI}}$ (including both desired Cartesian force and torque) by setting $\pmb{k_{FRI}}=\pmb{0}$ and $\pmb{F_{FRI}}=K(\pmb{x}^*-\pmb{x})$, getting a controller equal to (\ref{eqt:imp_control}) where $K$ is the stiffness matrix and the dynamics $ \pmb{f_{dyn}}$ and damping $\mathrm{diag}(\pmb{d_{FRI}})\pmb{v}$ are managed by the KUKA's internal controller.

In practice, due to noise in the demonstration from human and measurement uncertainty, averaging over a chosen number of time steps to compute $P$ in (\ref{eqt:polyhedron}) produces more stable results. To filter the noise, we chose to average over 20 time steps of original 100Hz measuring frequency, which meant sampling $P$~in~5Hz. We used manually estimated values of 20 degrees for $\eta$ and 10 degrees for $\xi$ in (\ref{eqt:epsilons})\footnote{Code available at www.irobotics.aalto.fi }. \changethree{These values provide a good starting point for any experiment; increasing the values causes longer segments to be detected, which can be valuable in certain use cases. Moreover, we set the stiffness values of the non-compliant axes $k$ to $200\frac{N}{m}$; this value depends on the robot, but it simply needs to make the robot be non-compliant. Damping was managed by the KUKAs interrnal controller, with ${d_{FRI}}=0.7$. further details on the effects of this choice are explained in Section~\ref{sec:desired}.    }

To evaluate the method for purely translational motion, we performed workpiece alignment on a similar valley setup as in \cite{suomalainen2017} consisting of two aluminium plates set on 45 degrees angle with the table. As expected, the generalized version presented in this paper produced similar results as the translation-only version presented in \cite{suomalainen2017} and thus the results are not included here for brevity. Since the rest of the results presented in this chapter require rotational compliance, they could not be completed with the translation-only algorithm from \cite{suomalainen2017} and thus there is no comparison between the results.

\begin{figure}[tbp]
\centering
\begin{subfigure}[b]{0.32\columnwidth}
	\centering
	\includegraphics[width=\columnwidth]{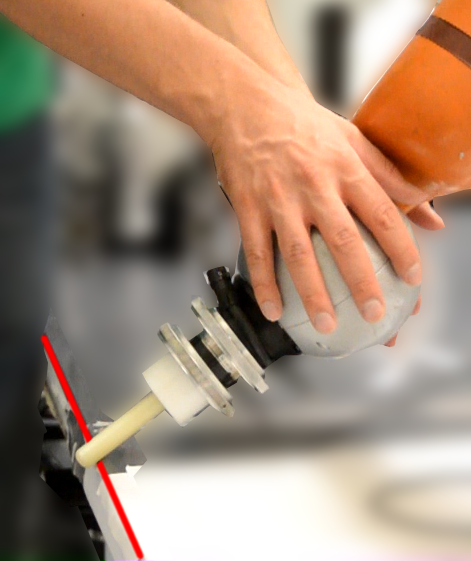}
\end{subfigure}
\begin{subfigure}[b]{0.32\columnwidth}
	\centering
	\includegraphics[width=\columnwidth]{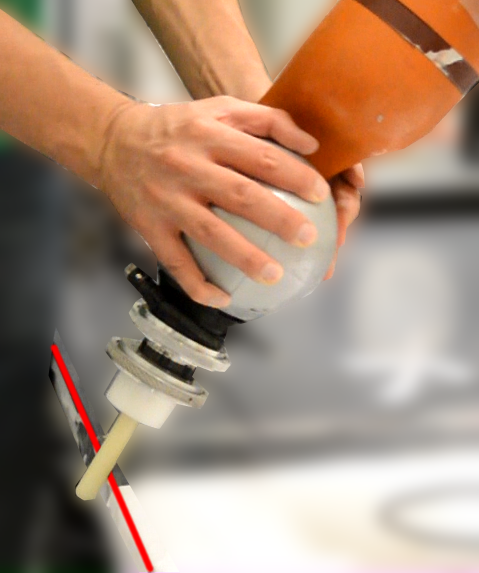}
\end{subfigure}
\begin{subfigure}[b]{0.32\columnwidth}
	\centering
	\includegraphics[width=\columnwidth]{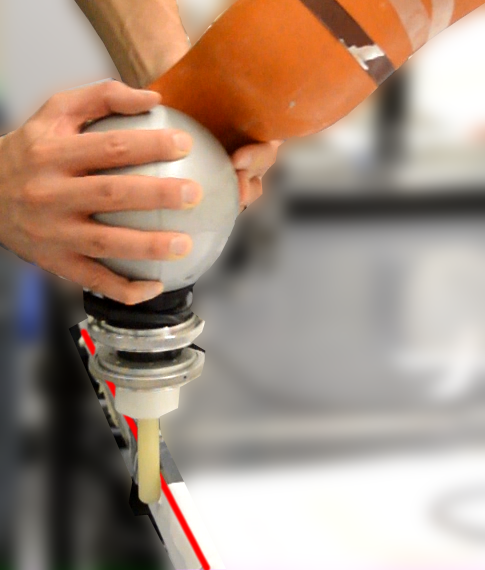}
\end{subfigure}
\caption{Screenshots from a demonstration of rotating the peg around the edge of the table, where the translational motion is caused by the contact forces. The edge of the table is highlighted in red.}
\label{fig:reuna}
\end{figure}

To evaluate the method for motions including rotation we performed four motions included in common contact tasks in households and industry. For each motion we first carried out one or more demonstrations and then let the robot perform the learned task. The peg-in-hole setup is a common contact problem where compliance is highly advantageous---with the setup shown in Fig.~\ref{fig:pih}, we analyzed whether the algorithm finds the correct parameters to slide the peg completely in when it starts from a wrong orientation but partly inside the hole. \changetwo{In this setup we also performed a comparison against a DMP with hand-tuned compliance.} The hose coupler setup shown in Fig.~\ref{fig:hose_coupler} presents another common aligning and interlocking task found in households and industry alike. With this setup we studied both the alignment phase with varying orientations as shown in Fig.~\ref{fig:hosealign} and the interlocking phase where the coupler is rotated to fix the parts together.  Finally, to study a case where rotations cause translations as explained in Section \ref{sec:full_compl}, we performed a motion where the peg is rotated around the edge of a table as shown in Fig.~\ref{fig:reuna}, a motion required whenever using a lever arm to increase the applied force.

We used an end-effector coordinate system defined at the wrist of the robot (the F/T sensor) in the experiments. However, the choice of the most suitable coordinate system is task-dependent. Whereas automatically choosing the coordinate system has been studied \cite{ureche2015task}, applying it in our context is outside the scope of this paper.

\subsection{Identification of desired direction of motion}
\label{sec:identification}
\begin{figure}[tbp]
\centering
\includegraphics[width=.4\columnwidth]{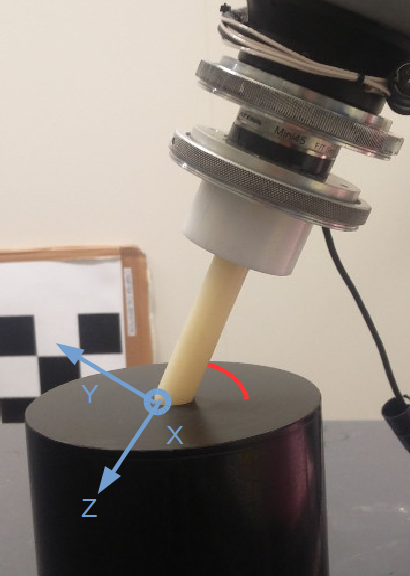}
\caption{Illustration (in red) of the angle measured in Fig.~\ref{fig:inliers}. The tool coordinate system used in the experiments is shown in cyan.}
\label{fig:pih_angle}
\end{figure}

Our goal was to study if 1) the inlier ratio check in (\ref{eqt:prop_vd}) can correctly identify whether $\pmb{\widehat{\omega}_{d}^*}$ and $\pmb{\hat{v}_{d}^*}$ are required and 2) if required, $\pmb{\widehat{\omega}_{d}^*}$ and $\pmb{\hat{v}_{d}^*}$ computed with Algorithm \ref{alg:vd} can reproduce the demonstrated motion. For this we used the peg-in-hole experiment setup, from which we recorded the angle between the peg and the plane as shown in \ref{fig:pih_angle}. From every 5 degree angle between 5 and 35 degrees, we performed 5 demonstrations by grasping the robot and leading the peg to the hole. 

\begin{figure*}[!t]
	\centering
	\includegraphics[width=\linewidth]{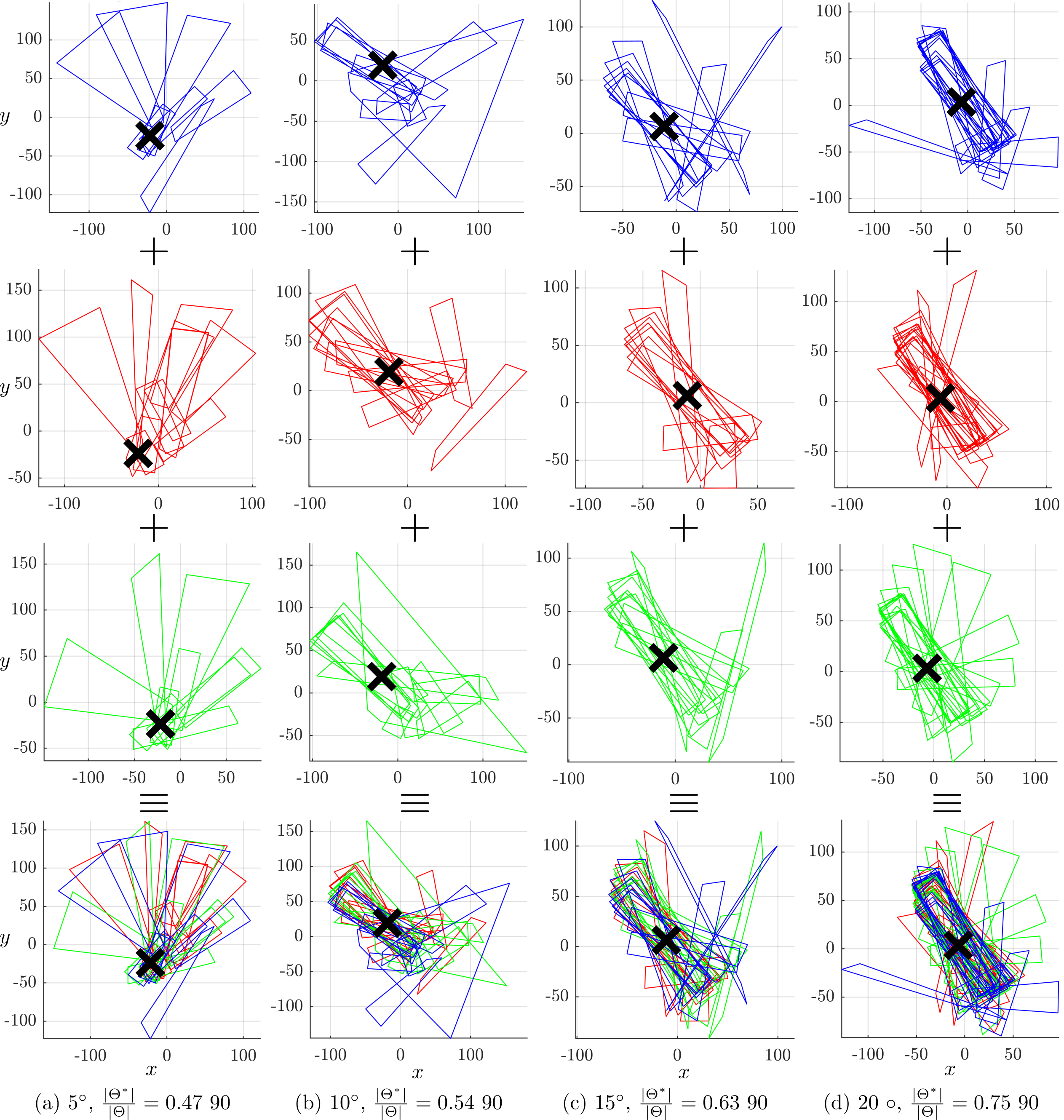}
	\caption{\changetwo{The rectangles used for choosing the desired direction (lines 9-21 of Algorithm \ref{alg:vd}) for rotations in the peg-in-hole task with three demonstrations (each color corresponds to one demonstration) such that each column (a)-(d) corresponds to demonstrations from a certain error angle (5-20 degrees) mentioned at the bottom of each column. The first three figures from the top on each column show the data from a single of the three demonstrations. and the bottom figure of each column shows all the demonstrations in a single figure. The black cross corresponds to the finally chosen desired direction. \changethree{The axes represent the coordinate system in the projection plane where the computations are made.} The inlier ratio $ \frac{\lvert\Theta^*\rvert}{\lvert\Theta\rvert}$ from (\ref{eqt:prop_vd}) is shown for each column. The figure is best seen in color.}} 
	\label{fig:inliers}
	\vspace{-1em}
\end{figure*}

A desired direction for translation was found for each angle approximately along the z-axis in tool coordinate system (Fig.~\ref{fig:pih_angle}). In this paper we chose to use three demonstrations for learning a desired direction -- a more thorough experiment of how the number of demonstrations affects the learning of desired direction was presented in \cite{suomalainen2017}, where we concluded that already one demonstration along each possible trajectory is enough to learn a valid $\pmb{\hat{v}_{d}^*}$. For finding the desired direction for rotation, Fig.~\ref{fig:inliers} shows 3 demonstrations with each starting angle of 5, 10, 15 and 20 degrees. It can be observed that the inlier ratio $ \frac{\lvert\Theta^*\rvert}{\lvert\Theta\rvert}$ steadily increases with the increase of the starting angle: \change{the rectangles over three demonstrations are well aligned in Fig.~\fa{\ref{fig:inliers}d} \iffalse\ref{fig:pih_rot20deg_rect}\fi, but in Fig.~\fa{\ref{fig:inliers}a} \iffalse\ref{fig:pih_rot5deg_rects}\fi less than half of the rectangles contribute to finding the intersection}. This corresponds to the fact that if the error angle (i.e. starting angle in this case, Fig.~\ref{fig:pih_angle}) is too large, a specific rotation needs to be introduced to complete the task. If, however, the error angle is low, it is enough to have compliance along the rotation together with a desired direction in translation. Our algorithm correctly captures this behaviour, and if the threshold $\zeta$ was set to 0.6, as would be natural for three demonstrations, $\pmb{\widehat{\omega}_{d}^*}$ would exist when error angle is 15 degrees or more. Naturally the demonstrations are not required to be started from strictly the same error angle- combining demonstrations with error angle 10 or less degrees showed similar results, as did combining demonstrations with error angle of 15 degrees or more. When $\pmb{\widehat{\omega}_{d}^*}$ was required, the direction was correctly identified along the rotation.

\begin{figure}[tbp]
\centering
\begin{subfigure}[b]{0.47\columnwidth}
	\centering
	\includegraphics[width=\columnwidth]{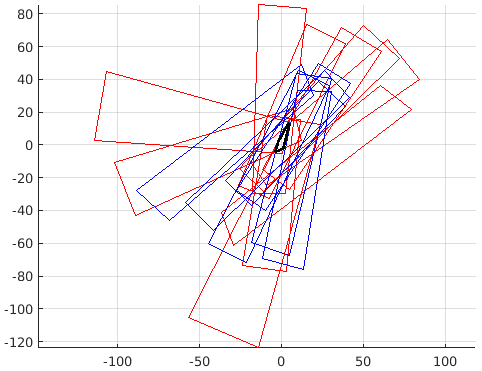}
	\caption{Hose-coupler alignment translations} \label{fig:hosealign_trans_rects}
\end{subfigure}
\begin{subfigure}[b]{0.47\columnwidth}
	\centering
	\includegraphics[width=\columnwidth]{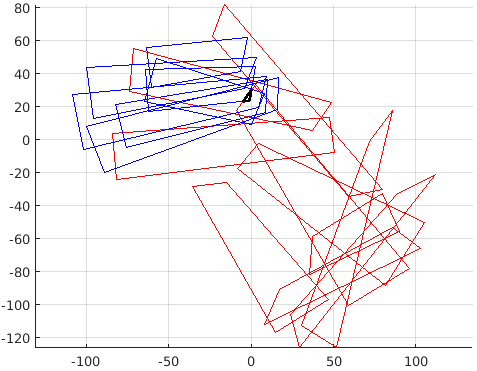}
	\caption{Hose-coupler alignment rotations} \label{fig:hosealign_rot_rects}
\end{subfigure}
\caption{The rectangles used for choosing the desired direction (lines 8-21 of Algorithm \ref{alg:vd}, \changetwo{or the projected bottom of the pyramid in Fig.~\ref{fig:single_pyramid}}) either for translations or rotations in the hose-coupler task. The red and blue colors indicate the two separate demonstrations of the task and the black rectangle is the intersection $\Phi$, i.e. the set of all desired directions in the angle coordinate system. \changetwo{The rotations from the two demonstrations are clearly apart, and thus the desired direction is discarded.}}
\label{fig:thetas}
\end{figure}

\begin{figure}[tbp]
\centering
\begin{subfigure}[b]{0.47\columnwidth}
	\centering
	\includegraphics[width=\columnwidth]{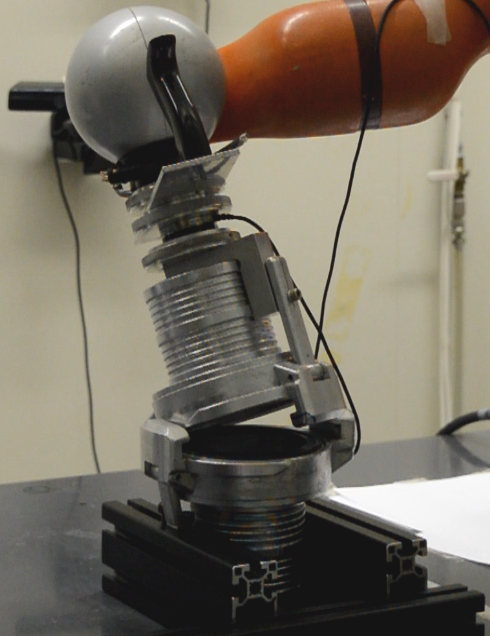}
\end{subfigure}
\begin{subfigure}[b]{0.47\columnwidth}
	\centering
	\includegraphics[width=\columnwidth]{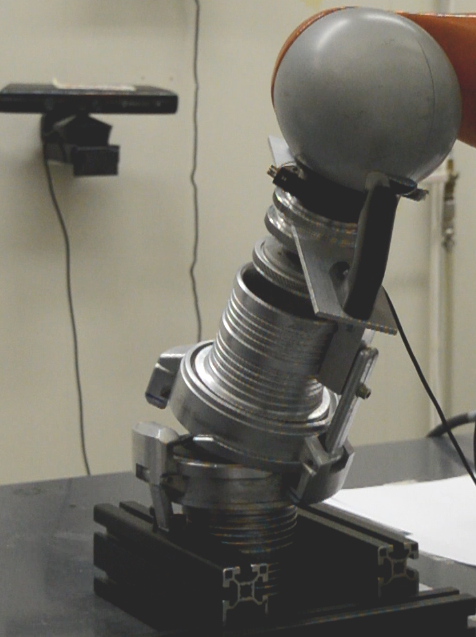}
\end{subfigure}
\caption{Two starting positions for demonstrations of the hose-coupler alignment task.}
\label{fig:hosealign}
\end{figure}

To study the identification of the desired direction in the hose-coupler alignment, two demonstration from starting positions shown in Fig.~\ref{fig:hosealign} were given. \change{The algorithm identified a desired translation direction $\pmb{\hat{v}_{d}^*}$, illustrated as the intersection shown as black polygon in \ref{fig:hosealign_trans_rects}. For the rotations, the maximal intersection covers poorly the demonstrations with inlier ratio of 0.41 as shown in Fig.~\ref{fig:hosealign_rot_rects}. Thus, the algorithm concluded correctly that there was no desired rotational direction $\pmb{\widehat{\omega}_{d}^*}$ and rotational compliance was sufficient to perform the rotational alignment which was demonstrated}. Also in the hose-coupler interlocking and peg-around-the-edge motions (Fig.~\ref{fig:reuna}), the desired directions were correctly identified to replicate the motions. We conclude that our method can correctly identify the desired direction for both rotations and translations, and motion in both can be correctly combined to reproduce tasks such as peg-in-hole with high error angle, which requires both rotational and translational motions.

\subsection{Learning axes of compliance}
\label{sec:dof}
\begin{figure}[tbp]
\centering
\begin{subfigure}[b]{0.47\columnwidth}
	\centering
	\includegraphics[width=\columnwidth]{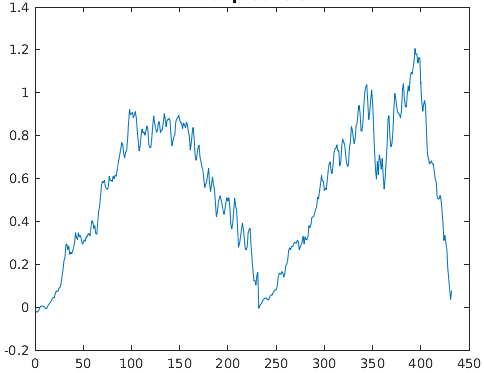}
	\caption{Translations}
	\label{fig:reuna_trans_dots}
\end{subfigure}
\begin{subfigure}[b]{0.47\columnwidth}
	\centering
	\includegraphics[width=\columnwidth]{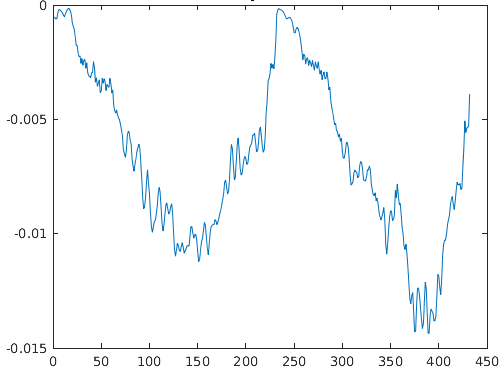}
	\caption{Rotations}
	\label{fig:reuna_rot_dots}
\end{subfigure}
\caption{The dot products between speed and force and between angular speed and torque over time on the peg-around-the-edge motion. } 
\label{fig:reuna_dots}
\end{figure}

Our goal was to study whether our method can find the number of compliant axes and their directions in $K_f$ and $K_o$ which, together with the desired directions $\pmb{\widehat{\omega}_{d}^*}$ and $\pmb{\hat{v}_{d}^*}$, can reproduce the demonstrated motion. In the peg-around-the-edge motion (Fig.~\ref{fig:reuna}), the demonstration was performed such that the demonstrator was only rotating the tool, and the translation at the wrist occurred due to coupling of the translational and rotational motions. Therefore it was recognized in (\ref{eqt:full_compl}) that the translations need to be 3-DOF compliant. To give an insight about this result, the dot products between speed and force and between angular speed and torque are plotted over time in Fig.~\ref{fig:reuna_dots}. It can be observed that with translations there is more work done by the environment than the demonstrator, since the curve stays on the positive semi-axis the whole time. The method correctly concluded that translations must be 3-DOF compliant in this motion. 

\begin{figure}[tbp]
\centering
\begin{subfigure}[b]{0.49\columnwidth}
	\centering
	\includegraphics[width=\columnwidth]{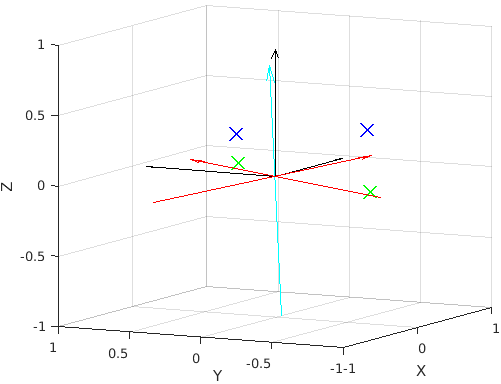}
	\caption{Translations}
	\label{fig:hosealign_trans_dof}
\end{subfigure}
\begin{subfigure}[b]{0.49\columnwidth}
	\centering
	\includegraphics[width=\columnwidth]{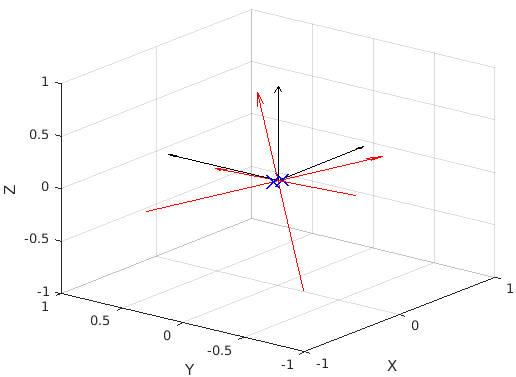}
	\caption{Rotations}
	\label{fig:hosealign_rot_dof}
\end{subfigure}
\caption{Illustrations of choosing the directions of compliant axes on the hose-coupler alignment experiment. The black arrows denote the world coordinate system, the red ones the eigenvectors U and the blue crosses the average motions of each demonstration, $\pmb{\overline{\psi}_{a}}$. In (a) $\pmb{\hat{v}_{d}^*}$ is plotted in cyan and the $\pmb{\bar{v}_{a}}$ with the component along $\pmb{\hat{v}_{d}^*}$ removed, as on line \ref{line:removedd} in Algorithm \ref{alg:compliant}, are plotted as green crosses. In both (a) and (b) 1 compliant axis is chosen} 
\label{fig:hosealign_dof}
\end{figure}

In the other case where most of the work is not done by the environment, the number of compliant axes and their directions must be detected individually. The directions of the compliant axes are directly the vectors of the chosen matrix $U_d$ from Algorithm \ref{alg:compliant}. Vectors from $U_3$, \change{i.e.~the candidate axes of compliance, are visualized in Fig.~\ref{fig:hosealign_dof} for the hose-coupler alignment task. In Fig.~\ref{fig:hosealign_trans_dof} a desired direction in translation $\pmb{\hat{v}_{d}^*}$ is detected, which overlaps as expected with one of the eigenvectors. Consequently, the component along $\pmb{\hat{v}_{d}^*}$ is removed from $\pmb{\bar{v}_{a}}$ (blue crosses) and they are projected onto the plane of the other eigenvectors (green crosses). The green crosses are far from the origin, meaning that compliance is required, but they fall along a single eigenvector, which leads to conclusion of a single compliant axis. In Fig.~\ref{fig:hosealign_rot_dof} there is no desired direction, and thus no projection is required. A line through the blue crosses indicates the direction of the single compliant axis detected by the algorithm.}



In the peg-in-hole experiments, at least one axis of compliance was detected for each error degree between 5 and 35. This is according to theory- without a desired direction, at least one compliant direction is required, whereas with a desired direction the compliant directions merely assist the motion. The difference is that whereas in 5-10 error degrees the first axis of compliance is found to approximately match the direction of motion, with higher error degrees the rotation motion is handled by $\pmb{\widehat{\omega}_{d}^*}$.  We conclude that the method correctly identified the compliant axes and their directions.

\subsection{Reproduction of motion}
Finally, to evaluate that the motions can be reproduced with the learned parameters, we performed the motions on all the aforementioned experiments. In \cite{suomalainen2017} we already showed that the learning of desired direction is robust by randomizing over multiple sets of demonstrations. Now we show the generalization capabilities in the peg-in-hole case- in particular, how much error can be tolerated with compliance alone, and when is actual rotation required.

In the peg-in-hole experiments, we first used parameters learned from all 5 demonstrations with 10 degrees of error. As shown in Fig.~\ref{fig:inliers}, no $\pmb{\widehat{\omega}_{d}^*}$ was found, but only $\pmb{\hat{v}_{d}^*}$ along z-axis (Fig.~\ref{fig:pih_angle}) moves the peg. Compliance is required and found both in rotations and translations- in translations it is found along y-axis and in rotations around x-axis. With these parameters we performed five reproduction attempts starting from ever 5 degree angle. The peg is successfully inserted with error angles 5-15 degrees. With an error angle of 20 degrees, friction prevents sliding and the motion is unsuccessful. This result is in par with the results of Section \ref{sec:identification}: demonstration of 15 degrees error is on the border regarding identification of desired direction for rotation, but this amount of error can still be handled with only desired direction in translation.  

For the cases where both $\pmb{\hat{v}_{d}^*}$ and $\pmb{\widehat{\omega}_{d}^*}$ were detected, $\pmb{\hat{v}_{d}^*}$ was again along z-axis but $\pmb{\widehat{\omega}_{d}^*}$, as expected, varied depending on the starting orientation of the tool. Nevertheless, for demonstrations recorded with 20 and 30 degrees error, reproduction was successful with the learned angle and lower angles but not on higher angles. These results are summarized in Table \ref{tab:own_reprods}. Thus we conclude that whenever the worst case scenario of error is demonstrated, the method can successfully interpolate to cases where the orientation error is smaller than in the demonstration. 

\changetwo{We also repeated the peg-in-hole experiment through a direct reproduction of a single demonstration for each angle using Cartesian-DMP \cite{abu2015adaptation} with  hand-tuned compliance in an impedance controller. The  results are shown in Table~\ref{tab:dmp_reprods}. Neither of the methods can reliably extrapolate to larger error angles, even though both methods succeed in this on occasions. However, the presented method can always manage lower error angles, whereas DMP with compliance struggles with these. This shows the main difference of the presented method to methods based on attractors; in many tasks simply carrying out a learned linear motion will result in success, whereas learning a certain trajectory is always to the vicinity of the trajectory. Nonetheless, attractor methods are useful in many tasks, and thus the choice of method should be task-dependent.  }

\begin{table*}[h]
    \begin{subtable}[h]{0.45\textwidth}
        \centering
        \begin{tabular}{@{}ccccccccc @{}}
  	{} & \multicolumn{8}{c}{\bf Reproduction angle} \\
         {} & {} & 5 & 10 & 15 & 20 & 25 & 30 & 35 \\ \cmidrule{2-9}
        {} & 10 & \color{green} \CheckmarkBold  & \color{green} \CheckmarkBold &  \color{green} \CheckmarkBold & \color{red} \textbf{X} & \color{red} \textbf{X} & \color{red} \textbf{X} & \color{red} \textbf{X}\\ \cmidrule{2-9}
        {\bf Demo angle} & 20 & \color{green} \CheckmarkBold  & \color{green} \CheckmarkBold &  \color{green} \CheckmarkBold & \color{green} \CheckmarkBold & \color{red} \textbf{X} & \color{red} \textbf{X} & \color{red} \textbf{X}\\ \cmidrule{2-9}
        {} & 30 & \color{green} \CheckmarkBold  & \color{green} \CheckmarkBold &  \color{green} \CheckmarkBold & \color{green} \CheckmarkBold & \color{green} \CheckmarkBold & \color{green} \CheckmarkBold & \color{red} \textbf{X}\\  \cmidrule{2-9}
  \end{tabular}
  \caption{Our method }
  \label{tab:own_reprods}
    \end{subtable}
    \hfill
    \begin{subtable}[h]{0.45\textwidth}
        \centering
        \begin{tabular}{@{}ccccccccc @{}}
  	{} & \multicolumn{8}{c}{\bf Reproduction angle} \\
         {} & {} & 5 & 10 & 15 & 20 & 25 & 30 & 35 \\ \cmidrule{2-9}
         
        {} & 10 & \color{red} \textbf{X}  & \color{green} \CheckmarkBold &  \color{green} \CheckmarkBold & \color{red} \textbf{X} & \color{red} \textbf{X} & \color{red} \textbf{X} & \color{red} \textbf{X}\\ \cmidrule{2-9}
        
        {\bf Demo angle} & 20 & \color{red} \textbf{X}  & \color{green} \CheckmarkBold &  \color{green} \CheckmarkBold & \color{green} \CheckmarkBold & \color{green} \CheckmarkBold & \color{red} \textbf{X} & \color{red} \textbf{X}\\ \cmidrule{2-9}
        
        {} & 30 & \color{red} \textbf{X}  & \color{red} \textbf{X} & \color{red} \textbf{X} & \color{green} \CheckmarkBold & \color{green} \CheckmarkBold & \color{green} \CheckmarkBold & \color{red} \textbf{X}\\  \cmidrule{2-9}
  \end{tabular}
  \caption{DMP with compliance }
  \label{tab:dmp_reprods}
     \end{subtable}
     \caption{ \changetwo{ A table summing up the results of reproduction experiments both with the method presented in this paper (a) and DMP with hand-tuned compliance (b). Each task was repeated 5 times, and the results were always failures only or successes only. Symbol {\color{green} \CheckmarkBold} marks success and {\color{red} \textbf{X}} marks failure. It can be seen that DMP has difficulties adjusting to smaller error angles than demonstrated, whereas neither method can reliably generalize to higher error angles.}}
     \label{tab:reprods}
\end{table*}

\begin{figure}[tbp]
\centering
\begin{subfigure}[b]{0.32\columnwidth}
	\centering
	\includegraphics[width=\columnwidth]{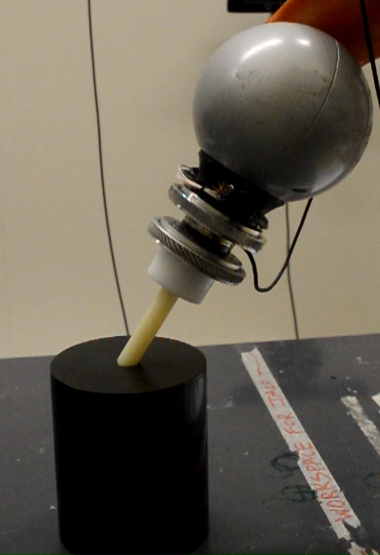}
	\label{fig:sup1}
\end{subfigure}
\begin{subfigure}[b]{0.32\columnwidth}
	\centering
	\includegraphics[width=\columnwidth]{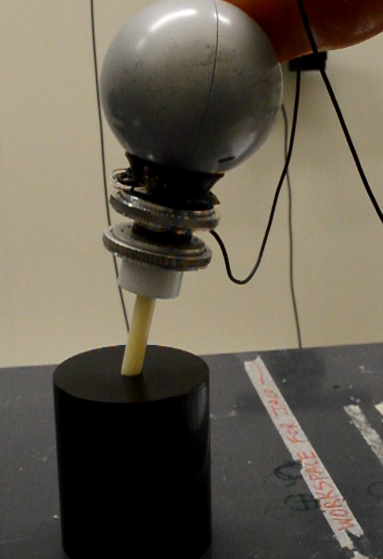}
	\label{fig:sup2}
\end{subfigure}
\begin{subfigure}[b]{0.32\columnwidth}
	\centering
	\includegraphics[width=\columnwidth]{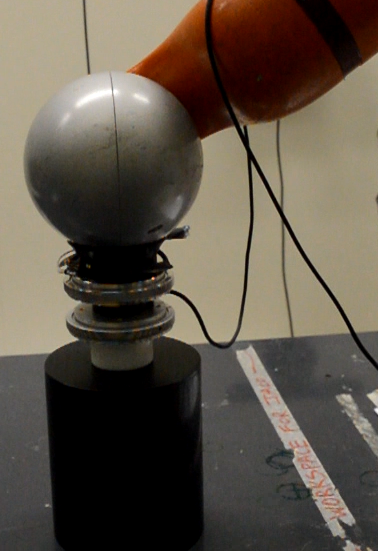}
	\label{fig:vino1}
\end{subfigure}
\caption{Screenshots from a reproduction video of the peg-in-hole motion. The motion starts from the leftmost picture, and the peg is rotated and pushed to the bottom. The peg has radius 16.5~mm, length 80~mm and a rounded tip, and the hole's radius is 0.25~mm more than the peg's.}
\label{fig:pih}
\end{figure}

In Fig.~\ref{fig:pih} are shown screenshots from a reproduction of the peg-in-hole reproduction with 30 degrees error. Our algorithm also successfully reproduced the demonstrated motion on the hose-coupler alignment, hose-coupler interlocking and peg-around-the-edge experiments. We conclude that the parameters our method learns from human demonstration can be used to perform the motions with an impedance controller primitive.

\subsection{\changetwo{Discussion}}
\label{discussion}

\changetwo{
In this paper we provided a general geometry-based approach to learn compliant motions from human demonstrations and adapt them to new situation within a region that we call it the convergence region. }

\changetwo{
The main strengths of our approach are:
\begin{enumerate}[a)]
    \item The ability to properly leverage contact when completing a task.
     \item The ability to converge to a goal in contact without exact localization of the robot w.r.t. the goal.
    \item No need for any prior information about the robot, the tool or the goal (as needed with \eg contact formations).
    \item Ability to start from new starting point within the convergence region (unseen in demos).
    \item No need for any numerical information about the change of the goal when extrapolating. 
    \item Ability to learn from a small number of demonstrations (application dependent, but typically two are enough).
    \item The ability to extrapolate to trajectories not shown during demonstration by leveraging contact.
\end{enumerate}
while the limitations are:
\begin{enumerate}[a)]
    \item Inability to perform nonlinear motions unless influenced by the environment.
    \item The stopping condition must be handled separately, either by force signal interpretation or other kind of tracking. Also there must be suitable geometry to facilitate stopping condition detection.
    \item No built-in mechanism for detecting sliding towards a wrong direction, away from the goal.
\end{enumerate} }

\changetwo{Experimentally (see Section \ref{EXPERIMENTS}), we evaluated the approach extensively using a real setup. The PiH task is used as an example of the practical usability of the proposed approach. However, the method can be applied to different applications, such as the presented hose coupler example, screwing, folding, or other assembly-like tasks that can be performed with linear motions in 6-D.}

\changethree{The stability of a manipulator in contact with the environment has to be handled carefully; the closed loop stability of a robot in contact with the environment depends on the environment’s compliance characteristics, the stiffness and damping. For the closed-loop system to be stable, the damping of the robot’s controller must be sufficient to dampen potential oscillations. We assume that the damping of the robot is above critical damping threshold corresponding to the maximum stiffness set for the proposed method, which depends heavily on the hardware and the application. Noting that the stiffness of the robot-environment system can not be larger than the robot’s own stiffness (only the robot’s compliance remains if the environment is perfectly stiff), the damping will be sufficient to dampen the oscillations of the robot in contact.
}

\changetwo{Besides the experimental comparison to DMP, some considerations can be made on how the presented method compares to others state-of-the-art LfD methods. 
Abu-dakka \etal\cite{abu2015adaptation} integrated iterative learning control (ILC) with DMPs in order to overcome the uncertainties due to the transformation of the acquired skills to the new starting pose. Their approach needs information about robot starting point and the convergence region is smaller than with the method presented in this paper. Moreover, unlike Abu-Dakka's approach, our method does not need to transfer the demonstration profiles to the new starting pose.}


\changetwo{SEDS \cite{khansari2011learning} was proposed to learn the parameters of the dynamic system to ensure that all motions follow closely the demonstrations while ultimately reaching in and stopping at the target. 
SEDS relies on GMM/GMR, but improves the EM learning strategy by incorporating stability constraints in
the likelihood optimization.
SEDS represents a global map which specifies instantly the correct direction for reaching the target, considering the current state of the robot, the target, and all the other objects in the robot’s working space. This makes SEDS state-based learning.
Although SEDS focused on stabilising movement trajectories, it did not stabilise impedance during interaction. However, SEDS has been extended in Khansari \etal\cite{khansari2014modeling} to learn motion trajectories while regulating the impedance during interaction and ensuring global stability.
Saying that, SEDS can learn much more complex attractor landscapes. }

\changetwo{To conclude, there are several major differences between the presented method and SEDS, which make them useful in different use cases. Firstly and most importantly, the presented method is not state-based; thus,  the presented method does not depend on the accuracy on knowing the coordinate transform between the robot end-effector and the goal. Secondly, related to this, the presented method does not have a specified target, and thus when the target is not geometrically different, SEDS or another method like it should be used. Thirdly, in contrast to our approach, SEDS needs much more data (demonstrations) for learning, particularly in high dimensions.}


Defining a \say{good} demonstration is a difficult task. There are existing attempts to measure \say{goodness} by coverage in case of free space motions \cite{sena2018teaching}, but for in-contact tasks there exists no current work. In the experiments we simply explained the demonstrators how an informative demonstrations should be performed---the development of a general metric of informativeness for compliant motion demonstrations is outside the scope of this paper and an interesting direction for future research.

\changetwo{The choice of the center of compliance is a prior design choice; the implications of this choice for peg-in-hole were researched in more detail in \cite{suomalainen2019improving} and concluded that the tooltip is the most suitable choice. It is, in general, beneficial to choose the center of compliance such that rotations are around it; however, whereas we did not specifically experiment this, there is no reason the presented method could not learn rotations around other points as well, such that rotations comprise of both rotations and translations. } 

\section{CONCLUSIONS AND FUTURE WORK}
We presented a method that can successfully learn and reproduce 6-D compliant motions from human demonstrations. The method finds a desired direction of motion which can be either pure translation, pure rotation or a combination of translation and rotation. Then it finds the compliant axes, both in translation and rotation, necessary to reproduce the motion. We found that compliance along rotation can compensate fairly significant errors in the angle. The exact angle depends heavily on the equipment, but in our setup the tolerance was fairly tight and a simple rounding of the tool's end created enough of convergence region to take advantage of compliance. Advantages of using compliance only include the ability to use the same controller in free space, as demonstrated with translations in \cite{suomalainen2017}. However, for cases where the angle is not due to error but due to instructions, we show that we can learn an active rotation as well. 

The method presented in this paper models an assembly task as a sequence of linear directions and compliances. Taking into account the physics of sliding in contact allows us to use intersection in the desired direction computations. Due to the use of intersection, it is easy to combine as many timesteps as required and thus the number of demonstrations or their ratio of lengths do not cause issues, in contrast to DMP which calculates the average over many demonstrations. Also since our method is programmed to perform the learned linear motion until physical constraints, our primitive generalizes to holes of different depth and chamfers of different length. Finally, not following a pre-specified force trajectory but instead using compliance to adapt to new situations makes our method more robust towards errors in the initial position of the motion. On the other hand, DMP-based methods would perform better in tasks which require non-linear motions in free space or motions where the final position of the motion is not physically constrained. 

A whole task would typically consist of a sequence of the primitives presented in this paper. Methods for sequencing primitives with linear dynamics is a common problem, for which various possible solutions have been presented \cite{kroemer2014learning,hagos2018seq}. The method presented in this paper is meant mainly for assembly tasks in situations where the coordinate transformations between the robot and the target are not accurately known and the use of vision is complicated. Such a situation arises in, for example, in small-to-medium size enterprises, where a robot must be included in an existing working environment and CAD models of the workpieces are not available.  

In \cite{suomalainen2017} the world coordinate system was used, while in this work we chose the tool coordinate system. Both coordinate systems have their advantages and disadvantages and the choice is task-dependent. A method to automatically choose the most suitable coordinate system would enhance the method's usability. 


\label{CONCLUSION}


\bibliographystyle{spmpsci}      
\bibliography{biblio}   

\begin{thebibliography}{10}
\providecommand{\url}[1]{{#1}}
\providecommand{\urlprefix}{URL }
\expandafter\ifx\csname urlstyle\endcsname\relax
  \providecommand{\doi}[1]{DOI~\discretionary{}{}{}#1}\else
  \providecommand{\doi}{DOI~\discretionary{}{}{}\begingroup
  \urlstyle{rm}\Url}\fi

\bibitem{abudakka2020Geometry}
Abu-Dakka, F.J., Kyrki, V.: Geometry-aware dynamic movement primitives.
\newblock In: IEEE International Conference on Robotics and Automation (ICRA),
  pp. 4421--4426. Paris, France (2020)

\bibitem{abu2015adaptation}
Abu-Dakka, F.J., Nemec, B., J{\o}rgensen, J.A., Savarimuthu, T.R., Kr{\"u}ger,
  N., Ude, A.: Adaptation of manipulation skills in physical contact with the
  environment to reference force profiles.
\newblock Autonomous Robots \textbf{39}(2), 199--217 (2015)

\bibitem{abu2018force}
Abu-Dakka, F.J., Rozo, L., Caldwell, D.G.: Force-based variable impedance
  learning for robotic manipulation.
\newblock Robotics and Autonomous Systems  (2018)

\bibitem{abudakka2020Variable}
Abu-Dakka, F.J., Saveriano, M.: Variable impedance control and learning -- a
  review.
\newblock Frontiers in Robotics and AI pp. 1--27 (2020)

\bibitem{ahmadzedah2017cylinders}
Ahmadzadeh, S.R., Rana, M.A., Chernova, S.: Generalized cylinders for learning,
  reproduction, generalization, and refinement of robot skills.
\newblock In: Robotics: Science and Systems, vol.~1 (2017)

\bibitem{argall2009survey}
Argall, B.D., Chernova, S., Veloso, M., Browning, B.: A survey of robot
  learning from demonstration.
\newblock Robotics and autonomous systems \textbf{57}(5), 469--483 (2009)

\bibitem{calinon2007learning}
Calinon, S., Guenter, F., Billard, A.: On learning, representing, and
  generalizing a task in a humanoid robot.
\newblock IEEE Transactions on Systems, Man, and Cybernetics, Part B
  (Cybernetics) \textbf{37}(2), 286--298 (2007)

\bibitem{denivsa2016learning}
Deni{\v{s}}a, M., Gams, A., Ude, A., Petri{\v{c}}, T.: Learning compliant
  movement primitives through demonstration and statistical generalization.
\newblock IEEE/ASME Transactions on Mechatronics \textbf{21}(5), 2581--2594
  (2016)

\bibitem{garkavi1964chebyshev}
Garkavi, A.L.: On the {Chebyshev} center and convex hull of a set.
\newblock Uspekhi Matematicheskikh Nauk \textbf{19}(6), 139--145 (1964)

\bibitem{guan2018efficient}
Guan, C., Vega-Brown, W., Roy, N.: Efficient planning for near-optimal
  compliant manipulation leveraging environmental contact.
\newblock In: Robotics and Automation (ICRA), 2018 IEEE International
  Conference on. IEEE (2018)

\bibitem{hagos2018seq}
Hagos, T., Suomalainen, M., Kyrki, V.: Estimation of phases for compliant
  motions.
\newblock In: Intelligent Robots and Systems (IROS 2018), IEEE/RSJ
  International Conference on. IEEE (2018, Accepted for publication.
  arXiv:1809.00686)

\bibitem{hogan1987stable}
Hogan, N.: Stable execution of contact tasks using impedance control.
\newblock In: Robotics and Automation. Proceedings. 1987 IEEE International
  Conference on, vol.~4, pp. 1047--1054. IEEE (1987)

\bibitem{huang2019kernelized}
Huang, Y., Rozo, L., Silv{\'e}rio, J., Caldwell, D.G.: Kernelized movement
  primitives.
\newblock The International Journal of Robotics Research \textbf{38}(7),
  833--852 (2019)

\bibitem{kalakrishnan2011learning}
Kalakrishnan, M., Righetti, L., Pastor, P., Schaal, S.: Learning force control
  policies for compliant manipulation.
\newblock In: 2011 IEEE/RSJ International Conference on Intelligent Robots and
  Systems, pp. 4639--4644. IEEE (2011)

\bibitem{khansari2011learning}
Khansari-Zadeh, S.M., Billard, A.: Learning stable nonlinear dynamical systems
  with gaussian mixture models.
\newblock IEEE Transactions on Robotics \textbf{27}(5), 943--957 (2011)

\bibitem{khansari2014modeling}
Khansari-Zadeh, S.M., Kronander, K., Billard, A.: Modeling robot discrete
  movements with state-varying stiffness and damping: A framework for
  integrated motion generation and impedance control.
\newblock Proceedings of Robotics: Science and Systems X (RSS 2014)
  \textbf{10}, 2014 (2014)

\bibitem{kramberger2017generalization}
Kramberger, A., Gams, A., Nemec, B., Chrysostomou, D., Madsen, O., Ude, A.:
  Generalization of orientation trajectories and force-torque profiles for
  robotic assembly.
\newblock Robotics and Autonomous Systems \textbf{98}, 333--346 (2017)

\bibitem{kramberger2016transfer}
Kramberger, A., Gams, A., Nemec, B., Schou, C., Chrysostomou, D., Madsen, O.,
  Ude, A.: Transfer of contact skills to new environmental conditions.
\newblock In: Humanoid Robots (Humanoids), 2016 IEEE-RAS 16th International
  Conference on, pp. 668--675. IEEE (2016)

\bibitem{kroemer2014learning}
Kroemer, O., Van~Hoof, H., Neumann, G., Peters, J.: Learning to predict phases
  of manipulation tasks as hidden states.
\newblock In: Robotics and Automation (ICRA), 2014 IEEE International
  Conference on, pp. 4009--4014. IEEE (2014)

\bibitem{lefebvre2005online}
Lefebvre, T., Bruyninckx, H., De~Schutter, J.: Online statistical model
  recognition and state estimation for autonomous compliant motion.
\newblock IEEE Transactions on Systems, Man, and Cybernetics, Part C
  (Applications and Reviews) \textbf{35}(1), 16--29 (2005)

\bibitem{mason1981compliance}
Mason, M.T.: Compliance and force control for computer controlled manipulators.
\newblock IEEE Transactions on Systems, Man, and Cybernetics \textbf{11}(6),
  418--432 (1981)

\bibitem{mukadam2020riemannian}
Mukadam, M., Cheng, C.A., Fox, D., Boots, B., Ratliff, N.: Riemannian motion
  policy fusion through learnable lyapunov function reshaping.
\newblock In: Conference on Robot Learning, pp. 204--219 (2020)

\bibitem{ohwovoriole1981extension}
Ohwovoriole, M., Roth, B.: An extension of screw theory.
\newblock Journal of mechanical design \textbf{103}(4), 725--735 (1981)

\bibitem{osa2018algorithmic}
Osa, T., Pajarinen, J., Neumann, G., Bagnell, J., Abbeel, P., Peters, J.: An
  algorithmic perspective on imitation learning.
\newblock Foundations and Trends in Robotics \textbf{7}(1-2), 1--179 (2018)

\bibitem{paraschos2013probabilistic}
Paraschos, A., Daniel, C., Peters, J.R., Neumann, G.: Probabilistic movement
  primitives.
\newblock In: Advances in neural information processing systems, pp. 2616--2624
  (2013)

\bibitem{peternel2015human}
Peternel, L., Petri{\v{c}}, T., Babi{\v{c}}, J.: Human-in-the-loop approach for
  teaching robot assembly tasks using impedance control interface.
\newblock In: Robotics and Automation (ICRA), 2015 IEEE International
  Conference on, pp. 1497--1502. IEEE (2015)

\bibitem{racca2016}
Racca, M., Pajarinen, J., Montebelli, A., Kyrki, V.: Learning in-contact
  control strategies from demonstration.
\newblock In: Intelligent Robots and Systems (IROS), 2016 IEEE/RSJ
  International Conference on, pp. 688--695. IEEE (2016)

\bibitem{reiner2014lat}
Reiner, B., Ertel, W., Posenauer, H., Schneider, M.: Lat: A simple learning
  from demonstration method.
\newblock In: Intelligent Robots and Systems (IROS 2014), 2014 IEEE/RSJ
  International Conference on, pp. 4436--4441. IEEE (2014)

\bibitem{rozo2013learning}
Rozo~Casta{\~n}eda, L., Calinon, S., Caldwell, D., Jimenez~Schlegl, P., Torras,
  C.: Learning collaborative impedance-based robot behaviors.
\newblock In: Proceedings of the Twenty-Seventh AAAI Conference on Artificial
  Intelligence, pp. 1422--1428 (2013)

\bibitem{schaal2006dynamic}
Schaal, S.: Dynamic movement primitives-a framework for motor control in humans
  and humanoid robotics.
\newblock In: Adaptive Motion of Animals and Machines, pp. 261--280. Springer
  (2006)

\bibitem{schimmels1991force}
Schimmels, J.M., Peshkin, M.A.: Force-assemblability: Insertion of a workpiece
  into a fixture guided by contact forces alone.
\newblock In: Robotics and Automation, 1991. Proceedings., 1991 IEEE
  International Conference on, pp. 1296--1301. IEEE (1991)

\bibitem{schreiber10}
Schreiber, G., Stemmer, A., Bischoff, R.: The fast research interface for the
  {KUKA} lightweight robot.
\newblock In: Proc. of the IEEE Workshop on Innovative Robot Control
  Architectures for Demanding (Research) Applications -- How to Modify and
  Enhance Commercial Controllers (ICRA 2010). IEEE (2010)

\bibitem{schwarz1978estimating}
Schwarz, G., et~al.: Estimating the dimension of a model.
\newblock The annals of statistics \textbf{6}(2), 461--464 (1978)

\bibitem{sena2018teaching}
Sena, A., Zhao, Y., Howard, M.J.: Teaching human teachers to teach robot
  learners.
\newblock In: 2018 IEEE International Conference on Robotics and Automation
  (ICRA), pp. 1--7. IEEE (2018)

\bibitem{stolt2015robotic}
Stolt, A.: On robotic assembly using contact force control and estimation.
\newblock Ph.D. thesis, Lund University (2015)

\bibitem{suomalainen2019improving}
Suomalainen, M., Calinon, S., Pignat, E., Kyrki, V.: Improving dual-arm
  assembly by master-slave compliance.
\newblock In: 2019 International Conference on Robotics and Automation (ICRA),
  pp. 8676--8682. IEEE (2019)

\bibitem{suomalainen2017}
Suomalainen, M., Kyrki, V.: A geometric approach for learning compliant motions
  from demonstration.
\newblock In: Humanoid Robots (Humanoids), 2017 IEEE-RAS 17th International
  Conference on, pp. 783--790. IEEE (2017)

\bibitem{ureche2015task}
Ureche, L., Umezawa, K., Nakamura, Y., Billard, A.: Task parameterization using
  continuous constraints extracted from human demonstrations.
\newblock IEEE Transactions on Robotics \textbf{31}(ARTICLE) (2015)

\end{thebibliography}

\end{document}